\documentclass[conference,compsoc]{IEEEtran}
\usepackage{tikz}
\usepackage{amsmath}
\usepackage{microtype}
\usepackage{tabularx}
\usepackage{url}
\usepackage{booktabs}
\usepackage{lipsum}
\usepackage{subcaption}

\definecolor{azure}{rgb}{0.0, 0.5, 1.0}

\usepackage{tikz}
\usepackage{pifont}
\newcommand{\xmark}{\ding{55}}%
\usepackage{amsmath,amsfonts,bm,amssymb,amsthm}
\usepackage{multirow}
\newtheorem{nullhypothesis}{Null Hypothesis}
\usepackage{enumitem}
\usepackage{microtype}
\usepackage{hyperref}
\usepackage{cleveref}

\begin{document}
\pagestyle{plain}

\date{}

\title{\Large \bf A Picture is Worth 500 Labels: A Case Study of Demographic Disparities in Local Machine Learning Models for Instagram and TikTok}

\author{\IEEEauthorblockN{Jack West\IEEEauthorrefmark{1}\textsuperscript{\textsection},
Lea Thiemt\IEEEauthorrefmark{2}\textsuperscript{\textsection}, Shimaa Ahmed\IEEEauthorrefmark{1}, Maggie Bartig\IEEEauthorrefmark{1}, Kassem Fawaz\IEEEauthorrefmark{1}, and Suman Banerjee\IEEEauthorrefmark{1}}
\IEEEauthorblockA{\IEEEauthorrefmark{1}University of Wisconsin-Madison\\
\IEEEauthorrefmark{2}Technical Unversity of Munich\\
Email: \IEEEauthorrefmark{1}\{jwwest,ahmed27,mbartig,kfawaz,suman.banerjee\}@wisc.edu,
\IEEEauthorrefmark{2}lea.thiemt@tum.de}}

\maketitle

\begin{abstract}
 Mobile apps have embraced user privacy by moving their data processing to the user's smartphone. Advanced machine learning (ML) models, such as vision models, can now locally analyze user images to extract insights that drive several functionalities. Capitalizing on this new processing model of locally analyzing user images, we analyze two popular social media apps, TikTok and Instagram, to reveal (1) what insights vision models in both apps infer about users from their image and video data and (2) whether these models exhibit performance disparities with respect to demographics. As vision models provide signals for sensitive technologies like age verification and facial recognition, understanding potential biases in these models is crucial for ensuring that users receive equitable and accurate services. 
 
We develop a novel method for capturing and evaluating ML tasks in mobile apps, overcoming challenges like code obfuscation, native code execution, and scalability. Our method comprises ML task detection, ML pipeline reconstruction, and ML performance assessment, specifically focusing on demographic disparities. We apply our methodology to TikTok and Instagram, revealing significant insights. For TikTok, we find issues in age and gender prediction accuracy, particularly for minors and Black individuals. In Instagram, our analysis uncovers demographic disparities in the extraction of over 500 visual concepts from images, with evidence of spurious correlations between demographic features and certain concepts. 

\end{abstract}

\begingroup\renewcommand\thefootnote{\textsection}
\footnotetext{Equal contribution.}
\begingroup\renewcommand\thefootnote{\IEEEauthorrefmark{2}}
\footnotetext{All work was performed at University of Wisconsin-Madison.}
\begingroup\renewcommand\thefootnote{1}

\section{Introduction}

Most mobile apps we use daily, such as TikTok and Instagram, are free, yet they provide advanced and unprecedented services, including social networking and messaging. Our data drives the financial model of the mobile ecosystem: mobile apps access our data, extract insights about us, and use the data to serve personalized services. This ubiquitous collection and analysis of user mobile data has led to a push towards privacy, fueled by user awareness~\cite{breitinger2020survey}, media revelations~\cite{nytimesAmazonsAlexa}, and privacy regulations~\cite{ncsl2020Consumer}. Mobile app vendors have reacted by moving their data processing to the user's smartphone. Instead of sending raw user data, such as images, to their cloud, algorithms running on the user's phone analyze the data and extract needed insights~\cite{protocolMachineLearning}. This trend has benefited dramatically from hardware and software processing advances, mainly in vision-based machine learning models that consume swaths of user images and videos on smartphones in real-time. Advanced vision models can now analyze every image and camera frame to extract insights that drive several app functionalities.

The local processing of user images offers a unique opportunity to study how mobile apps process user images. More importantly, we can investigate whether vision models in apps exhibit performance disparities with respect to demographics. This investigation is important because vision models provide signals for sensitive technologies, such as age verification~\cite{tiktokTikTokMake} and facial recognition systems. Understanding potential biases in these models is crucial for ensuring that diverse populations, especially from minority backgrounds, receive equitable and accurate services. Until recently, as data analysis happened mainly on the cloud, to which researchers do not have explicit access, research was limited in observing how ML models in apps process user data. For example, researchers have analyzed the distribution of advertisements on browsing apps~\cite{reardon201950} or have asked users to request their data from mobile service providers~\cite{wei2020twitter}, such as Twitter. Other works have investigated the biases in online APIs, such as face recognition~\cite{DBLP:Gendershades, nanda20umdfairness}. However, these studies are limited in scope, focusing only on a select number of APIs, not covering vision models currently in mobile products, and evaluating bias through datasets with limited semantic and demographic attributes variability.

\subsubsection*{Research Questions} We analyze the two most popular social media apps on Android: TikTok and Instagram. With over 1.2 billion users~\cite{searchenginejournalSocialMedia}, TikTok is one of the most popular social media platforms.
Users can upload their own short videos and interact with other users' content, e.g., by commenting or liking. Instagram is a social media app with over 2.1 billion users~\cite{demandsageInstagramStatistics}. Like TikTok, Instagram allows users to post short videos on the platform called Reels~\cite{instagramIntroducingInstagram}. Our analysis answers two questions: (1) \textit{what insights do vision models in TikTok and Instagram infer about users from their images and camera frames?} And (2) \textit{are there demographic disparities in the quality of the inferred insights?}

\subsubsection*{Challenges} Answering the first question, we have to analyze both apps, which is challenging because they are not open-sourced. While developers package their models within the apps and run them locally~\cite{sun2021mind}, there are three challenges in analyzing the models: obfuscation, native code execution, and scalability. First, apps apply code obfuscation techniques to hide variable and class names, making them less amenable to static analysis~\cite{graux2019obfuscated}. Besides, apps load several components dynamically at runtime, further complicating the static analysis task. Second, apps execute their machine learning algorithms in native code~\cite{sun2021mind}, for which dynamic analysis has little context about the task being executed. Third, scalability poses a significant challenge when monitoring applications due to the complexity of the mobile runtime environment, such as that of Android. Most applications execute tens of thousands of functions a second, and capturing each function call dynamically at runtime is not feasible.

Answering the first question allows us to identify the vision models in TikTok and Instagram along with their input data and output results. This, however, does not answer the question of the associated demographic disparities. Apps might harm users because of the inherent biases in the underlying vision models. These biases could have security and privacy repercussions, including misidentifying users~\cite{xiang2022being} or associating them with inaccurate concepts (see Table~\ref{tab:grey_background_concepts}). For example, an age prediction model that is less accurate for minority users puts them at a disadvantage when creating new accounts, if such a model is used for age verification~\cite{itmunchTikTokScans}. Similarly, children might bypass the age verification system allowing them to access the platform~\cite{cspanSocialMedia}. 

This task of analyzing demographic disparities in apps is challenging for two reasons. First, we have to feed the model a large dataset of images in real-time without having offline access to the model. Second, we need access to a large dataset of images that are correctly annotated and cover diverse demographic and semantic attributes, such as faces with different skin tones, hairstyles, or accessories. Existing datasets, such as FairFace~\cite{karkkainenfairface} and CelebA~\cite{zhang2020celeba}, cover a limited number of attributes: gender, age, and ethnicity.

\subsubsection*{Method}
We develop a novel methodology to dynamically capture and evaluate machine learning (ML) tasks in mobile apps. This methodology consists of three layers that overcome the first set of challenges: ML task detection, ML pipeline reconstruction, and ML performance assessment. The first layer presents a dynamic instrumentation approach that logs function invocations without overloading the device. We utilize these logs to identify entry points for ML-relevant functionality (Section~\ref{sec:detection_layer}). The second layer combines static and dynamic analysis to rebuild call stacks, trace function calls, and complete the ML execution pipeline (Section~\ref{sec:pipeline_layer}). The final layer evaluates the performance of the ML models by injecting custom datasets that are demographically diverse into the identified pipeline (Section~\ref{sec:assessment_layer}).

\begin{figure}[t]
    \centering
    \includegraphics[width=0.6\columnwidth]{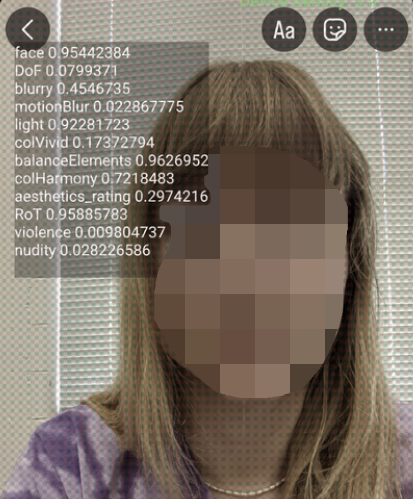}
    \caption{Some concept scores that Instagram extracts from an image that is about to be posted. The extraction happens whenever a user chooses an image from their device. These concept scores are from the Instagram debug layer, which is not enabled by default.}
    \label{fig:insta_screen}
\end{figure}

\subsubsection*{Findings}

We apply our methodology to both Instagram and TikTok, where we identify their use of different vision models locally on the device. These vision models consume camera frames and images that the user is about to post (before actually posting them), and they extract a set of concepts for each input. In particular, we find that TikTok continuously analyzes camera frames to detect faces and predict the corresponding age and gender. We trace the outputs of these models to an encrypted file that is saved on the device. We then analyze the performance of these models using the FairFace dataset~\cite{karkkainenfairface}. Our evaluation reveals that age prediction fails drastically for individuals below 19 and that gender prediction is problematic for black individuals. The detailed results are in Section~\ref{sec:tiktok}.

As for Instagram, we find a model that extracts more than 500 concepts from each image the user is about to post and from camera frames. Fig.~\ref{fig:insta_screen} shows a sample of the concepts extracted by Instagram. We then build a custom dataset that contains synthetic images of faces belonging to diverse demographics, with different semantic attributes corresponding to some of the concepts from Instagram. Our analysis of this model reveals significant demographic disparities in its performance, particularly for face-related concepts. Also, we find evidence of spurious correlations in the model performance, where some non-facial concepts are correlated with (have significantly higher scores for) images associated with particular demographic groups. For example, the ``great\_wall\_of\_china" concept is correlated with Asian women, and ``nudity" is correlated with White men. More detailed results are in Section~\ref{sec:instagram}.

\subsubsection*{Contributions}

In summary, our work is the first to investigate and evaluate the performance of models deployed by Instagram and TikTok. Our contributions are as follows:

\begin{itemize}[leftmargin=*]
\item We develop a novel methodology to dynamically capture and evaluate ML tasks in mobile apps. This methodology consists of three layers: ML task detection, ML pipeline reconstruction, and ML performance assessment.\footnotemark
 
\item We build a dataset to assess model performance on different demographics. Our dataset has two components: (1) public datasets and (2) synthetic data. The public datasets were chosen to measure model performance for age and sex. Whereas, our synthetic dataset evaluates targeted facial features within each demographic.\footnotemark

\item Using our dataset, we evaluate the effectiveness of age verification. We find that age verification, if done using the model deployed by TikTok, is less effective for younger demographics.

\item We analyze demographic correlations within Instagram's model. Since Instagram assigns over 500 different concepts to a single image, we went beyond facial features and evaluated underlying biases of concepts to racial demographics (see Table~\ref{tab:grey_background_concepts}). We found that certain demographics are correlated with specific concepts more than others. 
\end{itemize}

\footnotetext{All resources, code, and datasets can be found here \url{https://github.com/wi-pi/500-labels-resources}.}

\section{Background}

Before delving into the details of our work, we provide background on Android OS and ML development. 

\subsection{Android Background}

\begin{figure}[t]
    \centering
    \includegraphics[width=\columnwidth]{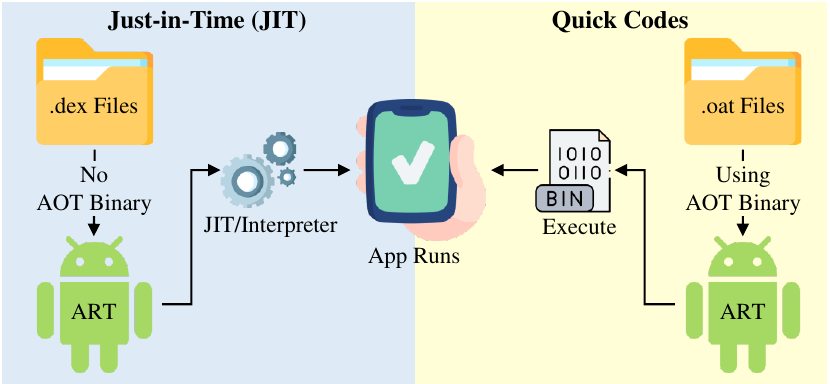}
    \caption{The Android Runtime (ART) executes the app's Ahead Of Time (AOT) compiled binaries if available. Otherwise, the ART passes the DEX files to the Just-In-Time (JIT) compiler or the Interpreter. Adapted from \cite{androidImplementingJustIntime}.}
    \label{fig:jit_compile}
\end{figure}

Android is a Linux-based operating system designed for mobile devices.
Android executes apps called Android Packages~(APKs), which contain assets, local files, resources, and executable code. There are two types of executable environments: Java and native. The operating system treats both differently at runtime~\cite{bleier2023ahead}.

Developers mainly write code in Kotlin or Java, and their language-specific compilers compile both languages into \textit{.class} files~\cite{bleier2023ahead}. Once compiled, an obfuscator then processes the \textit{.class} files to create DEX files.
DEX files treat Java and Kotlin the same way, essentially representing each as Java bytecode. On the other hand,  native code is directly compiled into a standard ARM64 ELF binary.
Using the compiled code, developers build the APK and upload it onto the operating system~\cite{bleier2023ahead}.

\subsubsection*{Android Runtime}
The Android Runtime~(ART) loads all APKs on the device.
Upon opening the APK, it loads all DEX files and the natively compiled code. Since 2019, the ART has used Ahead Of Time (AOT) compilation alongside Just-in-Time~(JIT) compilation~\cite{bleier2023ahead}. AOT compilation creates \textit{.oat} files, which are essentially ELF binaries with additional OAT-specific sections~\cite{liefprojectAndroidFormats} from the DEX files. The system does not compile some DEX files directly into an \textit{.oat} file.~\cite{androidImplementingJustIntime}. The JIT or Java interpreter handles the remaining DEX files at runtime. Figure~\ref{fig:jit_compile} shows the code execution paths in Android. 

On the other hand, native code is directly executed on the CPU in their compiled format.
For an APK to execute native code, it must use the Java Native Interface~(JNI)~\cite{xue2017malton}.
Native libraries are typically introduced to an Android app to increase performance~\cite{lin2011benchmark} or implement third-party code~\cite{almanee2021too}.
This feature is especially useful for computationally expensive tasks. Because of the computational needs, JNI calls commonly facilitate ML execution~\cite{sun2021mind}. 

\subsubsection*{Java Native Interface}
A JNI call executes in three phases:  call the JNI linking function, translate function arguments into native arguments, and execute the native function.
In Java, apps create links to JNI calls through the \textit{native} keyword~\cite {androidTipsAndroid}.
Android's Native Development Kit~(NDK) links \textit{native} functions to the shared library~\cite{androidConceptsAndroid}.
JNI functions, at the native level, have two arguments: the \textit{JavaVM} and the \textit{JNIEnv}.
The JavaVM variable allows apps to have a JavaVM in native code.
A JavaVM can interact, create, and destroy Java objects.
Whereas the JNIEnv performs local thread storage. 
These low-level objects prepare data for transmission or translate data for native usage~\cite{oracleFunctions}.
Then, the native code executes as it would on any Linux system.

There is, however, another type of JNI function that bears mentioning: JNI trampolines.
JNI trampolines determine their native address dynamically.
The operating system catches all trampoline calls and prepares the call for its actual execution point~\cite{android_source_code}.
The \textit{artQuickGenericJniTrampoline} retrieves the actual native code and then executes it on the CPU directly.
The return values occur asynchronously~\cite{android_source_code} in the \textit{GenericJniMethodEnd} function.

\subsubsection*{Quick Codes}
At the OS level, all JNI and AOT calls are \textit{quick codes} within the Android Runtime (ART).
Quick codes are a C++ object that represents the executable code for a function.
Quick codes are directly executed by the CPU at runtime.
When executing the code, the ART passes the arguments to the CPU as a raw C pointer.
A high-level C++ object represents the return value of the executing function.
Executing the function sets the value of the return object.
To access the arguments, the ART uses a \textit{shorty}; a shorty is a character array that indicates the return value type and the types of the arguments.
Later in the work, we describe how we interpret the \textit{shorty} to retrieve the function arguments and return value.

\subsubsection*{Native Callbacks}
Native code also communicates with Java code asynchronously using \textit{callbacks}~\cite{onespanUnderstandingCallback}.
Message objects, transmitted using a \textit{callback}, facilitate the communication.
A handler in the Java layer receives the messages.
All applications implement callbacks uniquely; however, applications commonly use the \textit{Android.os.message} object.
Current works trace a callback function by the name of the callback object~\cite{fourtounis2020identifying}.
When the objects are obfuscated, determining what the callback object is becomes complicated~\cite{fourtounis2020identifying}.

\subsubsection*{ROMs}

Tools like Frida~\cite{fridaFridaWorldclass} require root access to function properly.
When working with Android, acquiring root differs depending on the Android OS.
There are two distinct categories for an Android OS: custom ROMs and production ROMs.
A custom ROM is an Android OS derivative that a user installs.
Manufacturers install production ROMs, which are closed-source.

Root access is normally attainable with a custom ROM, because it is common for production ROMs disallow root access entirely.
However, it is possible to attain root on production ROMs,~\cite{xdaforumsGuideNovember}, which requires the phone to be \textit{OEM unlocked}.
In such a case, we gain root access by first downloading the production ROM's boot image and overwriting it with Magisk~\cite{githubGitHubTopjohnwuMagisk}. We then replace and boot into the new boot image.
After initialization, Magisk's Android app will have root privileges, install the `su' binary, and deploy Frida on the device.

Applications detect if they are on a custom ROM by utilizing Google's integrity API~\cite{kozSafetyNetGoogles}.
The integrity API allows app developers to gain insights into the device their app is currently running on.
Therefore, it is crucial to evaluate an app on both a production ROM and our custom ROM.
In this paper, we perform our valuation and analysis on both a custom ROM and a production ROM.

\subsection{ML Development}

There are two main categories of ML development on Android: through third-party libraries or in-house libraries~\cite{sun2021mind,ren2024demistify}.
Third-party libraries are open source and/or have public APIs~(Pytorch, TensorFlowLite, etc.).
Private groups develop in-house libraries that are inaccessible to the public.
Our work deals with both types of libraries, namely Pytorch~\cite{pytorchPyTorch}, a third-party library, and Pitaya~\cite{pitayaCitation}, an in-house library.  

\subsubsection{Pytorch}
Pytorch is an open-source library commonly used on mobile devices~\cite{sun2021mind}.
Instagram uses Pytorch's API calls to perform their ML calls.
Pytorch has a publicly available Java API that apps utilize to perform ML tasks.

\subsubsection{Pitaya}
Pitaya is Bytedance's own ML library~\cite{pitayaCitation}.
Pitaya is privately deployed to Bytedance's applications.
Pitaya consists of both native and Java components.
Our work will focus on their native library \textit{libbytenn.so}, which controls their ML engine.
Native executions control TikTok's computer vision ML suite. Java code never directly calls any ML API while the camera of the application is active. This implies that prior works~\cite{sun2021mind,ren2024demistify}, that do not trace native code, cannot detect Bytedance's apps ML behavior.

\section{Related Work}

In the following, we review recent work related to investigating machine learning and app analysis on Android. We also discuss some studies on bias measurements for machine learning. 

\subsection{Android Machine Learning Studies}

As on-device ML is becoming popular in Android apps~\cite{xu2019first, almeida2021smart, sun2021mind, deng2022understanding}, researchers developed methods to access models in order to study model protection and robustness.
One common approach to insert adversarial examples into on-device models is instrumenting standard Java APIs of ML frameworks~\cite{huang2021robustness, huang2022smart, cao2023cheating, deng2022understanding}.
To extract models from the apps, Sun et al.~\cite{sun2021mind} additionally dumped the memory location into which the model is loaded.
Ren et al.~\cite{ren2024demistify} developed a framework that reconstructs the Java calls associated with ML starting from exported JNI functions of native libraries. 
While existing approaches can deal with ML invocations initiated from the Java level, they cannot trace or reconstruct ML which is solely executed in the native layer without explicit triggers from Java functions.
Moreover, previous works rely on the fact that certain Java or native functions associated with ML are not obfuscated \cite{deng2022understanding, huang2022smart} or have some form of meaningful descriptions in the code \cite{ren2024demistify}. 
Although this works for a majority of apps \cite{sun2021mind, ren2024demistify}, it introduces limitations for apps that are heavily obfuscated, use closed-source custom-made libraries, or run ML in the native layer.
We investigated the works in~\cite{sun2021mind,huang2022smart,deng2022understanding,cao2023cheating}, which perform a keywords search or execution flow tracing within Java code. These works cannot detect TikTok's computer vision models for three reasons: (1) no Java code is executed during the ML task, (2) all ML tasks happen in native libraries, and (3) the models' format is closed source.
Furthermore, we observe that no previous study of ML on Android analyzes the performance disparity of the deployed models in production. We aim to close this gap with our case study of Instagram and TikTok ML models disparities and spurious correlations.

\subsection{Android Application Analysis}

Android malware analysis frameworks tackle the problem of detecting malware at scale.
While different from our work, their frameworks accomplish similar goals.
Isohara et al. ~\cite{isohara2011kernel}, instrumented the Android kernel for low-level malware analysis.
They built a logging framework in Logcat and captured logs within the instrumented device.
Their malware detection looks for root abuse within the kernel syscalls. Our work builds a similar logging infrastructure but instruments the ART rather than the Kernel.
Martinelli et al. ~\cite{martinelli2016find} also designed a framework for malware detection that has a similar goal to the work above but uses n-gram pattern matching on kernel syscalls to understand an app's behavior.
Keyes et al. ~\cite{keyes2021entroplyzer} used Frida for the dynamic classification of malware. They showed they were able to classify malicious activity at 98\%.
Our work uses Frida in a similar manner, but we use a physical Android device, and we also perform static analysis.
Samhi et al. ~\cite{samhi2022jucify} presented Jucify, a tool to statically analyze native code within an Android application.
The tool examines native call graphs but is limited by the tools used to extract the call graphs.
Our methodology targets specific areas within the natives, allowing us to rebuild important call graphs fully.

\subsection{Machine Learning Model Bias Studies}

Machine Learning models of various modalities, such as vision, speech, text, and other forms of input data, have been known to exhibit performance disparities that disproportionately impact certain demographic groups~\cite{mehrabi2021survey, barocas2017fairness, barocas2016big}. In the context of face images, Buolamwini and Gebru studied face recognition datasets and classifiers~\cite{buolamwini2018gender} and found that these classifiers exhibit demographic disparities.
This early work reported the impact of non-diverse training datasets on ML fairness.
Building on this, Raji and Buolamwini studied the impact of publicly releasing bias results on the performance of subsequent face recognition APIs~\cite{raji2019actionable}. Grother et al.~\cite{grother2019face} highlighted, in a NIST report, national origin and gender disparities of facial recognition algorithms. More recently, studies have shifted towards mitigation strategies for fairness. Selbst et al. ~\cite{selbst2019fairness} discussed the limitations of technical solutions for AI fairness in complex sociotechnical systems.
Stanovsky et al.~\cite{savoldi2021gender} focused on gender bias in machine translation, offering a framework applicable to vision models. Other works explored synthetic data to diversify the training and benchmarking datasets~\cite{colbois2021use, rosenberg2023unbiased}. However, generative models are found to exhibit bias trends similar to the existing natural datasets~\cite{seshadri2023bias, smith2023balancing, munoz2023uncovering}. Rosenberg et al.~\cite{rosenberg2023unbiased} provide a framework, using diffusion models, to generate synthetic face images that are demographically balanced with control over the facial semantic features. In our work, we assess the performance disparity of vision ML models deployed at the scale of billions of users, on commercial Android devices, unlike all other works that study open-source models or closed-box APIs with obscured knowledge of how these APIs manifest into the end-user experience. For a thorough assessment, we follow the work of Rosenberg et al.~\cite{rosenberg2023unbiased} to curate a demographically diverse dataset of face images with multiple semantic labels that align with the apps ML models labels.

\begingroup\renewcommand\thefootnote{2}
\section{ML Analysis Methodology}
\label{sec:methodology}

The methodology we develop aims to identify, reconstruct, and assess the performance of ML tasks in Android applications.
We design our approach to work directly on physical devices, which creates a realistic test environment and can directly use hardware features like the camera.
We develop our methodology through the following three layers, as evident in Figure~\ref{fig:framework_overview}.

\begin{figure}
    \centering
    \includegraphics[width=\columnwidth]{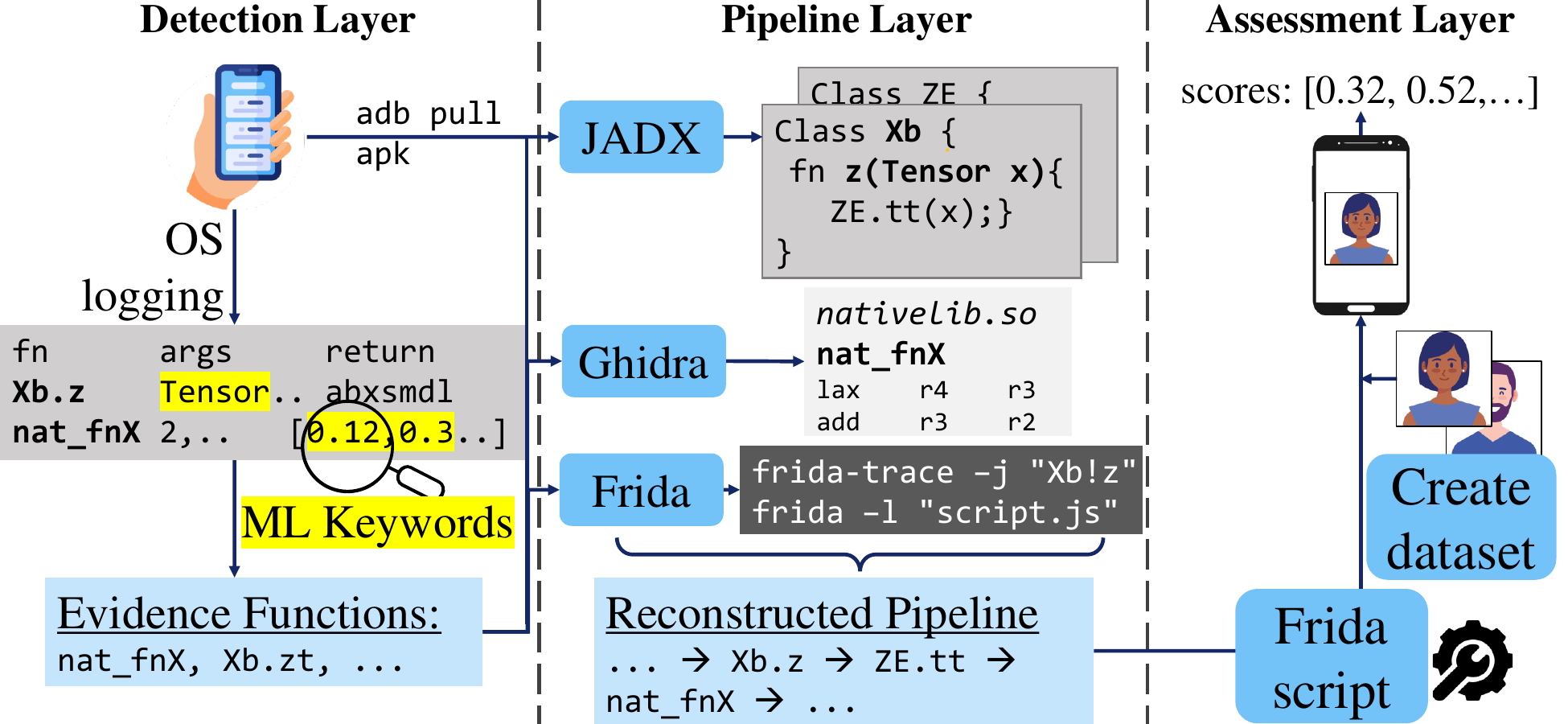}
    \caption{Our approach consists of three layers. The goal is to detect ML tasks in an app, reconstruct the ML pipeline, and assess the model. }
    \label{fig:framework_overview}
\end{figure}

\subsubsection*{ML Detection Layer}
The ML detection layer allows us to search for ML activity during runtime throughout an app's execution. We achieve this objective by instrumenting the ART to log Java/Kotlin and native function calls. We also log function arguments and return values from the lower layers of the ART. Using keywords from prior works~\cite{xu2019first,sun2021mind}, we flag the parts of the logged data where the keywords appear as \textit{evidence}. Once we have found the evidence, we can reconstruct the ML pipeline in the second layer. \smallskip 

\noindent
\textit{Key Challenge:}
Applications execute thousands of function calls a second, which quickly slows down the OS if it logs all of them. To address this challenge, we develop a targeted logging system that captures function calls in strategic locations within the ART.\smallskip 

\subsubsection*{Pipeline Reconstruction Layer}
In the second layer, we reconstruct an app's ML pipeline using the first layer's evidence as a starting point. We then design a novel hybrid approach based on combining static and dynamic analysis to rebuild call stacks,  further trace function calls, and complete the ML execution pipeline. The result is a layout of Java and native functions starting at the input to a model and ending at the points where the model results are saved or deleted. \smallskip

\noindent
\textit{Key Challenge:}
The core challenge is reconstructing the ML pipeline, which crosses the Java and Native code. 
Further, we observed some apps, such as TikTok, utilize non-exported functions in their ML tasks, which execute without explicit JNI calls. Existing frameworks typically focus on analyzing Java-level calls or executions through JNI calls, rendering them less effective against apps utilizing non-exported functions.\smallskip

\subsubsection*{ML Performance Assessment Layer}
The final layer evaluates the performance of ML models, focusing on demographic disparities. In particular, we use the pipeline identified in the previous layer to build a Frida script, which automatically injects data inputs into the model. The script then captures the model's output, which shows the confidence scores for the respective output labels. By obtaining the output scores for a diverse set of inputs, we can assess the demographic disparities in the model performance. Note that we assess the model by interacting with it directly during the app execution. We do not steal the model or reconstruct its weights, which infringes on the intellectual property of the app vendor. 
\smallskip

\noindent
\textit{Key Challenge:}
Analyzing demographic disparities in app-based ML models is challenging. First, we have to feed the model a large dataset of inputs (images in our case) in real-time without having offline access. Second, we need access to a large dataset of inputs that are correctly annotated and cover diverse demographic and semantic attributes, such as faces with different skin tones, hairstyles, or accessories. Existing datasets, such as FairFace~\cite{karkkainenfairface} and CelebA~\cite{zhang2020celeba}, cover a limited number of attributes: gender, age, and ethnicity.\smallskip

\subsection{ML Detection Layer}
\label{sec:detection_layer}

Our key insight for identifying ML is that when ML tasks are executed, they should leave evidence in natural language throughout function execution logs. While previous works follow the same reasoning and use indicative function names~\cite{sun2021mind} or comments~\cite{ren2024demistify} to identify ML, we find that this alone is not sufficient to capture ML behavior for every app, in particular, TikTok. Code obfuscation and the use of non-exported methods make it hard to identify evidence of ML execution from just function names and comments. However, we observe that apps serialize ML results, e.g., in JSON format, before further processing. We also observe that some apps use meaningful strings in input and output data structures. We leverage this observation to search for ML evidence in arguments to functions and return values.

\subsubsection{Existing Approaches}
Other approaches perform static analysis on decompiled Java and native libraries to find ML-related functions \cite{xu2019first,ren2024demistify} and use these to create scripts for further dynamic analysis.
While these approaches can cover a broad range of apps, they are not sufficient to detect relevant ML execution in every app. 
The reason is that performing static analysis to detect ML exhibits two shortcomings.
First, there can be ML functions in native libraries that are not exported and thus do not have a searchable name.
These ML functions may be called by other native functions instead of through explicit JNI calls from the Java layer. 
Second, we do not know when the app executes the statically detected functions or if it executes these functions at all.
This means that static analysis approaches may generate false positives if the ML functions are inside unused code.

\subsubsection{Methodology}

At a high level, we detect evidence for ML execution by logging the targeted function calls within an app. 
One approach to achieve this objective is through hooking into all of the functions with Frida.
This approach, however, is not feasible because of the large number of functions in an app, which would overload this kind of monitoring. 
Therefore, we build a more scalable method by instrumenting the ART to log function calls at runtime together with their passed arguments and return values.

To implement our logging system, we instrument the methods in the ART that are responsible for executing function invocations.
By managing function invocations, these ART methods handle invoked function parameters and return values.
We edit the ART methods to log each invoked function name, parameters, and return values. Additionally, we log the corresponding process id, thread id, and timestamp. The two ART methods of interest are the one that executes quick codes and the other one that executes JNI trampolines. Instrumenting these two methods ensures we can log all targeted function invocations without having to instrument each function invocation separately. Now, we discuss how we log quick code executions and JNI trampolines. \smallskip

\begin{table}[t]

    \centering
        \caption{ The \textit{shorty} characters indicate the types of the arguments and return values of quick code functions executed through the ART. To log the quick code execution, we created the mapping of shorty characters to data types by reading the ARM64 assembly source code provided by LineageOS. This may be different for other ROMs. }
    \begin{tabularx}{\columnwidth}{c c c c}
    
        \toprule
        \textbf{Character} & \textbf{Bytes} & \textbf{Description} & \textbf{Return Value Function} \\
        \midrule
        V & 4 & Void & GetV() \\
        Z & 4 & Boolean & GetZ() \\
        B & 4 & Byte & GetB() \\
        S & 4 & Short & GetS() \\
        C & 4 & Char & GetC() \\
        I & 4 & Int & GetI() \\
        J & 8 & Long & GetJ() \\
        F & 4 & Float & GetF() \\
        D & 8 & Double & GetD() \\
        L & 4 & Pointer & GetL() \\
        \bottomrule
    \end{tabularx}
    \label{tab:character_description}
\end{table}

\noindent
\textbf{{Logging Quick Code Execution.}}
The method for quick code execution takes an array for the arguments and a string called \textit{shorty}, which contains information about the return values and the type of the arguments. The first character of the shorty string always corresponds to the return value. The return values are stored in a results object. As these methods are not documented, we went through the Android OS code to reverse engineer them and understand how to instrument them. We found that each character in a \textit{shorty} represents a type and byte size as described in Table~\ref{tab:character_description}.
For instance, the shorty \textit{IIJ} indicates that the function returns an integer and takes two arguments, which are an integer and a long.

Therefore, we parse the shorty strings of each invoked function to determine the type and size of each argument. This allows us to correctly log the values and iterate over the array with the appropriate step size. We note that for non-static functions, the first argument is the object's \textit{this} pointer, which we skip by stepping over the first four bytes of the argument array. The result of an invoked function is stored in a special native object. To retrieve its value, we call a specific function tied to that object, which is determined by the first character in the shorty string. For example, we call \texttt{result->GetI();} to retrieve the value from the result object if the first character in the shorty string is ``I."

The values we read for arguments and return values are raw bytes.
As we know their type, we can correctly cast these bytes to log them in a readable format. An exception is the ``L" type which is a pointer to an object or array of primitives. The ART method does not receive the size of this object because it assumes the receiving function knows its size. Thus, instead of casting  ``L" values to a certain type, we log a fixed constant of 500 bytes. \smallskip

\noindent
\textbf{{Logging JNI trampoline execution.}}
To log the arguments of the function that executes JNI trampolines, we instrument the \textit{BuildGenericJniFrameVisitor} object, which is called to prepare the arguments for the JNI call.
As this object steps through the arguments, we log each argument similarly to the logging quick code execution. The return values are stored in a results object through the function \textit{GenericJniMethodEnd}.
We log these values by calling a function on the results object based on the object's type. \smallskip

\noindent
\textbf{{Detecting ML function candidates.}}
The ART instrumentation logs function invocations with arguments and return values. We then perform a keyword search to find the potential ML-related functions. We use the same keywords for detecting ML as prior work~\cite{sun2021mind, xu2019first}. Additionally, our logs allow us to search for serialized results of ML tasks. For instance, our logging system allows us to search for evidence of probability values calculated by a model,  which could be displayed as an array of values within $[0,1]$. The only way to obtain these logs is through running the app, forcing the ML execution, and logging input arguments and output results. It is difficult to obtain them through static analysis as they are often not hard coded in the APK files. The outcome of this layer is a set of candidate functions that execute ML tasks.

\subsection{ML Pipeline Layer}
\label{sec:pipeline_layer}

Our objective in the ML pipeline layer is to rebuild the program slice that executes the ML task, starting from the input to the model until the output of the model's inference values. We call this program slice the ML pipeline. We need this ML pipeline in order to execute the model with custom inputs and capture its output as we describe in the ML assessment layer.
It is important to ensure the reconstructed pipeline is complete such that we know that the inference values are calculated solely on the given input.
For instance, an app that determines a user's age through an ML model might not only use the user's image data but also in-app behavior, like their interaction with content. 
With the fully reconstructed pipeline, we can determine the values and data types passed to the model. To reconstruct the ML execution on the Java/Kotlin layer, we decompile the APK and use static analysis starting at the detected functions from the ML detection layer to trace the execution flow.
If the functions of interest run in the native layer or there is cross-language communication~\cite{gui2022cross,wei2018jn} (between Java and native), we additionally perform static analysis on the native libraries. We amend our analysis with dynamic analysis, using Frida, to trace and confirm the data flow and generate call stacks.

\subsubsection{Existing Approaches}
As explained in section \ref{sec:detection_layer}, other approaches commonly use static keyword analysis to detect ML functions within Java and native libraries. However, the reconstructed ML pipelines are only in the Java layer or use JNI functions, which are exported from native libraries. This strategy fails to rebuild an ML pipeline if the ML execution happens solely inside the native functions.
The first reason is that the ML function might not be discovered in the first place as it is not labeled with a name when it is not exported (we address this issue in section \ref{sec:detection_layer}).
Second, the function calls in a native library can depend on addresses that are dynamically calculated at runtime and cannot be determined statically.

\subsubsection{Methodology}
The ML-related functions we identify in the first layer can be at an arbitrary point in the execution chain.
Therefore, we inspect functions both prior to and after the call. First, to assist in the recovery of previous functions, we use Frida to hook into the given functions and display the call stacks in both Java and native code. We note that the functions in the reconstructed pipeline may have obfuscated names. Since these obfuscations can vary between different devices and app versions, we adapt the Frida scripts per APK instance. Then, we reconstruct the complete ML pipeline by exploring both decompiled Java functions and native libraries.

We use JADX to decompile the APK and statically analyze the Java code by following previous approaches~\cite{ren2024demistify,sun2021mind,xu2019first,lin2023fa3,xue2018ndroid,amin2019androshield}. One caveat here is that prior approaches decompile APKs downloaded directly from an app store, instead of retrieving it from the device. Such approaches would miss important functionality contained in the split APKs. So, we make sure to analyze the split APKs as well. To rebuild ML pipelines on the native layer, we perform static analysis by using Ghidra to disassemble and decompile the native libraries.

Since ML execution may also be performed in non-exported functions, we trace ML execution throughout the native libraries using a hybrid static-dynamic approach. 
Tracing execution flows within non-exported functions is challenging to perform through static analysis because of jump instructions to addresses, which are determined at runtime and are nondeterministic. To determine at which address such an instruction continues, we use dynamic analysis to monitor the execution and find the jump destination.

Toward that end, we first take the address inside the native library of this jump instruction. Then, we use Frida to hook into this address dynamically. This approach allows us, once the instruction is executed, to obtain the address where the execution continues. While this address is not constant due to memory randomization, it still shows us the byte offset within the native library, which is static. Using this byte offset, we can then continue statically tracing the native function execution. If we find multiple destination addresses for the jump, we follow each execution flow. We disregard those flows that do not contain ML as these are not important to reconstruct the ML pipeline.

\subsection{ML Assessment Layer}
\label{sec:assessment_layer}

The final layer of our methodology assesses the ML models through the reconstructed pipelines. In particular, we load datasets into ML models and capture the outputs while the application is running. This task requires generating realistic and correctly labeled images that cover a range of demographic and semantic
attributes. Further, we need to execute the reconstructed pipeline based on how the model is invoked in the app. If the model invocation requires a user to interact with the application, e.g., a user uploads some data, we directly execute the pipeline and call this \textit{internal injection}. On the other hand, an app might continuously or periodically execute a model. For instance, an app might constantly compute inference scores on frames in a camera video stream. In this case, we do not need to trigger the inference calls since it is sufficient to place the input and intercept the outputs, which we call \textit{external injection}.

\subsubsection{Existing Approaches}
There exist various datasets which other approaches use to assess the performance of vision models.
These datasets often show the general accuracy of models with respect to the label. Such datasets are not designed to evaluate the bias of a model with respect to a certain feature. Even if the datasets contain diverse images, they do not contain the features that the model in an app considers.
For instance, the model in a given app may calculate the probability that a person wears a necklace. 
Existing datasets would not contain enough sample images of individuals wearing a necklace, with diverse demographic attributes, to sufficiently assess the model's performance.

\subsubsection{Methodology}

From the ML pipeline we reconstructed in the previous layer, we know the ML model and its output labels. 
To create a meaningful assessment of the model, we create a dataset tailored to the features that the model extracts. As will be evident later, we assess vision models for faces. As such, we utilize and extend the framework by Rosenberg et al.~\cite{rosenberg2023unbiased} to synthesize images of faces belonging to diverse demographics. We apply different semantic attributes to the generated faces using concepts of interest from Instagram. We then manually annotate the generated dataset to ensure the images are not distorted and their ground truth labels are correct.  

We design both internal and external data injection such that we can efficiently evaluate the model on datasets. Note that in both cases, we utilize the app's default behavior as a vehicle to execute the ML pipeline. Our approach has the benefit of ensuring the ecological validity of the model analysis. Also, it does not require accessing or reconstructing the model weights, which has ethical and legal implications.  
\smallskip

\noindent
\textbf{Internal Injection.} We execute the complete pipeline which we recovered in the previous layer. Specifically, we use the recovered sequence of functions to build a Frida script that injects custom data into the model, initiates the inference calls, and then captures the output. From the ML pipeline, we know into which data type to cast the input images and how to parse the model outputs.\smallskip

\noindent
\textbf{External Injection.}
For external injection, we also use a Frida script to capture the output. However, since the application triggers the inference call itself, we inject the dataset outside of the device. For example, if the app directly feeds the model camera frames, then we display images on a screen in front of the device.

\section{TikTok}
\label{sec:tiktok}

TikTok offers video filters that users can use to change their appearance or add features to their videos. Implementing such filters is commonly performed through ML models which operate on a user's live camera feed. We find that there is one computer vision model that TikTok employs on the user's camera feed which, among others, predicts the sex and age of the user. In the following, we show how we extract this model, and we report its performance. 

\subsection{Analysis}

First, we discuss how we applied our methodology from Section~\ref{sec:methodology} to TikTok by providing the implementation details for the three layers. 

\subsubsection{ML Detection}
We focus our analysis on the computer vision model deployed by TikTok.
By searching for probability values in $[0,1]$, we detect logging messages~(\textit{MessageCenter.postMessage}) which display a face count, bounding box, age, and value named ``boy\_prob". We hypothesize that the last value describes the probability that the image contains the face of a male person. We thus refer to this variable as sex.
These logging messages show up always when the camera is active, even when no filter is selected.

\subsubsection{Pipeline Reconstruction}

\begin{figure}
    \centering
    \includegraphics[width=0.5\textwidth]{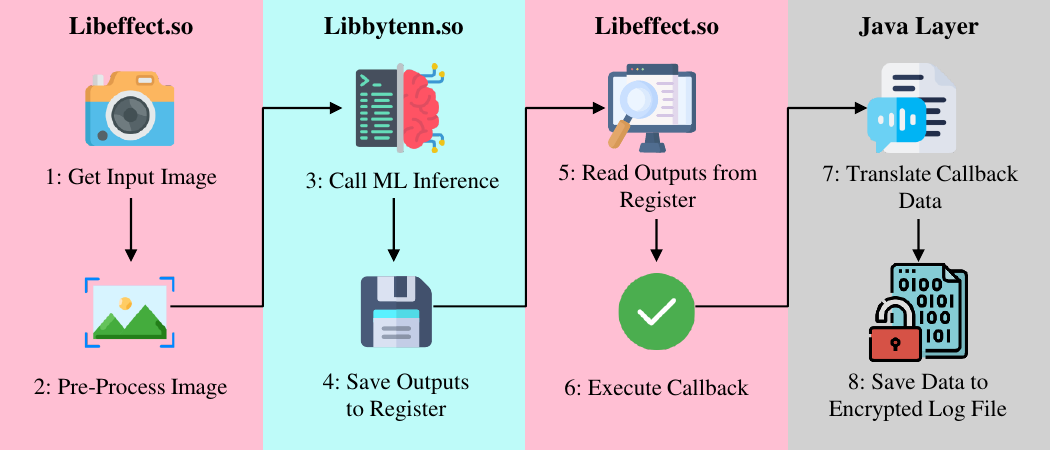}
    \caption[]{ In TikTok's ML pipeline, the native library \textit{libeffect.so} handles image data and directly calls the native ML functions in \textit{libbytenn.so}. Through a Java layer callback, the data is then saved in a log file. Note that this ML pipeline differs from the common case where apps exeucte native (ML) functions through JNI calls from the Java layer. }
    \label{fig:tiktok_flow}
\end{figure}

We use these logging messages as a starting point to reconstruct the complete ML pipeline as shown in Figure~\ref{fig:tiktok_flow}.
Through a string search of ``boy\_prob", we find that the logging function comes from the \textit{libeffect.so} library which is responsible for applying face filters.
This library has a native hook into the camera feed~(1), applies pre-processing on each frame~(2) and passes them to a computer vision model in \textit{libbytenn.so} which performs an inference call on the image data~(3) and saves the output to a thread safe register~(4).   
Note that it is a native library which directly calls the native ML functions.
This differs from most Android apps which perform ML calls through a JNI call from Java~\cite{sun2021mind,ren2024demistify}.
Back in the face filter library, the logging function reads the inference values from the register~(5) to build the message and execute the callback~(6) which we catch in the OS logs.
We highlight that the logging function is a native callback which acts as an asynchronous message handler.
TikTok uses their native callbacks to send the information from the native to the Java layer.
The logging function is not an exported function, meaning it has no external entry point, and there are no comments describing this function. 
This means that automated frameworks likely overlook its existence~\cite{ren2024demistify,sun2021mind}.
The native callback is received by a Java function~(7) which saves the data to an encrypted log file on the device~(8).
In summary, we discovered the ML pipeline inside the native libraries through a logging function which is not directly related to ML.
Through our pipeline reconstruction we are able to confidently connect it to the ML process.

\subsubsection{Performance Assessment}

In the pipeline reconstruction, we found that the inference call on images happens continuously on the camera feed without a user's explicit action.
Therefore, we do not need to rebuild and execute the whole ML pipeline.
Instead, we utilize the existing ML execution and inject the images externally by displaying them to the phone's camera through a display.
We capture the model outputs by hooking into the native callback function containing the ``boy\_prob" message with Frida.
For each image, we then intercept the message function which displays the inference values for sex and age.

\subsection{Evaluation}

Our focus in the ML assessment layer is on the sex and age values that TikTok's model predicts. The dataset we use in our performance measurement is the FairFace dataset~\cite{karkkainenfairface}. The FairFace dataset includes over 108,501 images that represent a diverse range of ages, genders, and seven different racial groups. We sample 44,903 images from the dataset where TikTok only detected a single face in the image. Each image in the dataset is labeled with an age range, an ethnicity, and a sex.

Our evaluation environment is a Pixel 4 on a production build on which we installed TikTok version 32.1.5 through the Google PlayStore.
We externally inject images by displaying them on a 1080p screen  within a dark isolated room using a Python script.
That same Python script hooks into the \textit{postMessage} function and captures the output using Frida's Python API.
TikTok also estimates the ``face\_count" within the image.
If the number of faces within an image is greater than 1, or if TikTok detects no face, we discard that datapoint. 
We run the experiment for two weeks and we collect over 40,000 different data points.
To evaluate the performance disparities, we define four null hypothesis groups:

\begin{nullhypothesis}\label{nullhyp:age_sex}
    The per-age distributions of  sex scores from TikTok are identical.
\end{nullhypothesis}

\begin{nullhypothesis}\label{nullhyp:eth_sex}
    The per-ethnicity distributions of sex accuracy from TikTok are identical.
\end{nullhypothesis}

\begin{nullhypothesis}\label{nullhyp:age_age}
     The per-age distributions of age scores from TikTok fall within the corresponding age bin.
\end{nullhypothesis}

\begin{nullhypothesis}\label{nullhyp:eth_age}
     The per-ethnicity distributions of age scores from TikTok are identical.
\end{nullhypothesis}

\begin{figure}[t]
    \centering
    \includegraphics[width=0.5\textwidth]{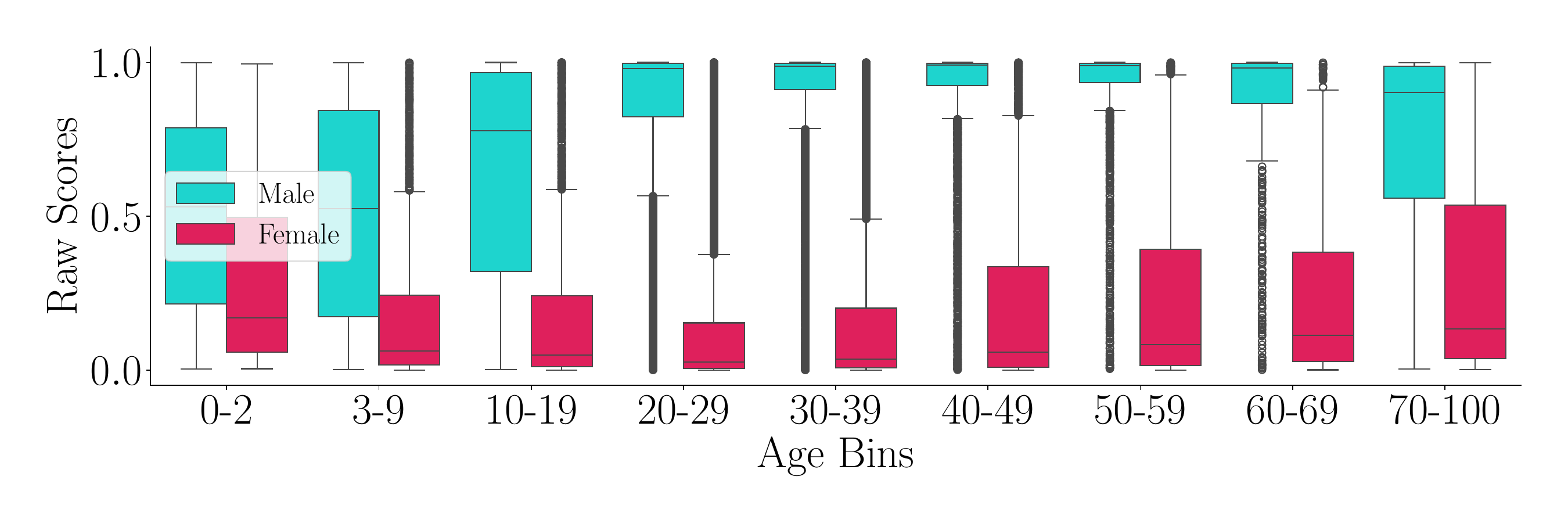}
    \caption[]{ TikTok's ML model assigns a score to a user's image which is close to 0 if the predicted sex is female and 1 if it is male. Analyzing the predicted sex (y-axis) grouped by age bin and labeled sex (x-axis), we observe that the model performs poorly for younger individuals. In older age bins, the model is more confident in predicting the sex of male individuals.}\footnote{The colors of this figure deliberately match TikTok's logo colors.}
    \label{fig:tiktok_age_gender}
\end{figure}

\footnotetext{\hypertarget{footnote2}{The} colors of this figure deliberately match TikTok's logo colors.}

\begin{figure}[t]
    \centering
    \includegraphics[width=0.5\textwidth]{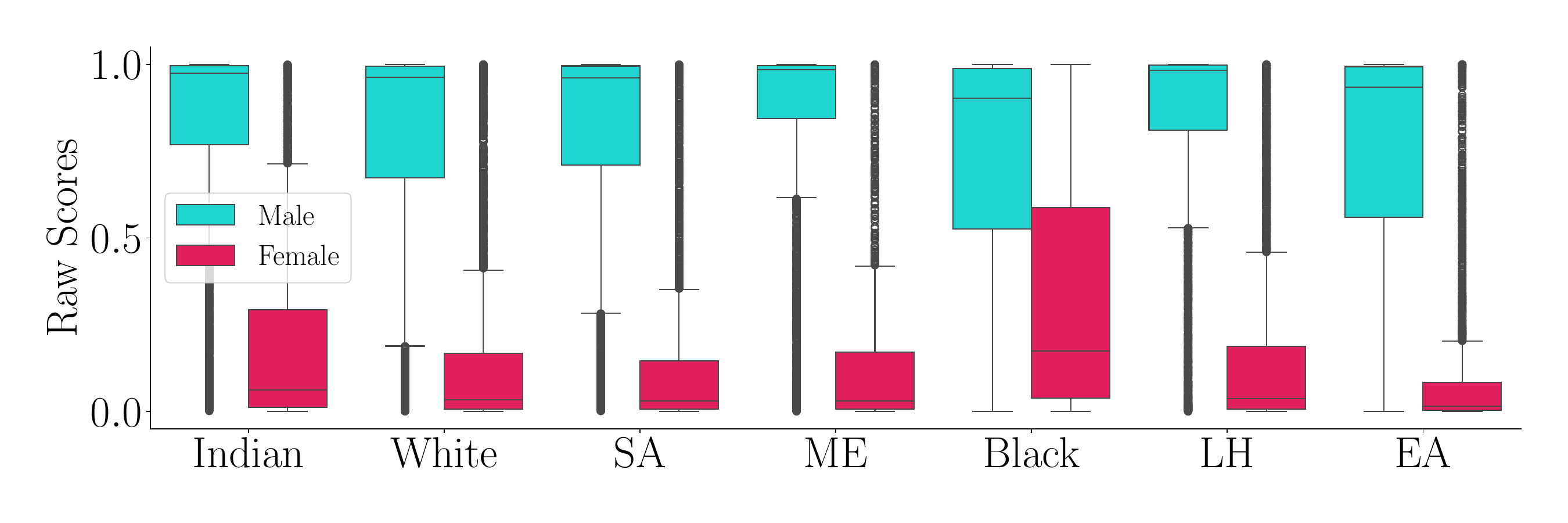}
    \caption[]{ On the x-axis, we abbreviated Middle Eastern to ME, Southeast Asian to SA, Latino Hispanic to LH, and East Asian to EA. Evaluating the distribution of TikTok's predicted sex scores (y-axis) grouped by ethnicity (x-axis), we find that the model is most confident in predictions for Southeast and East Asian individuals and least confident for Black individuals.\hyperlink{footnote2}{\textsuperscript{2}}} 
    \label{fig:tiktok_gender_ethnicity}
\end{figure}

\begin{figure}[t]
    \centering
    \includegraphics[width=0.5\textwidth]{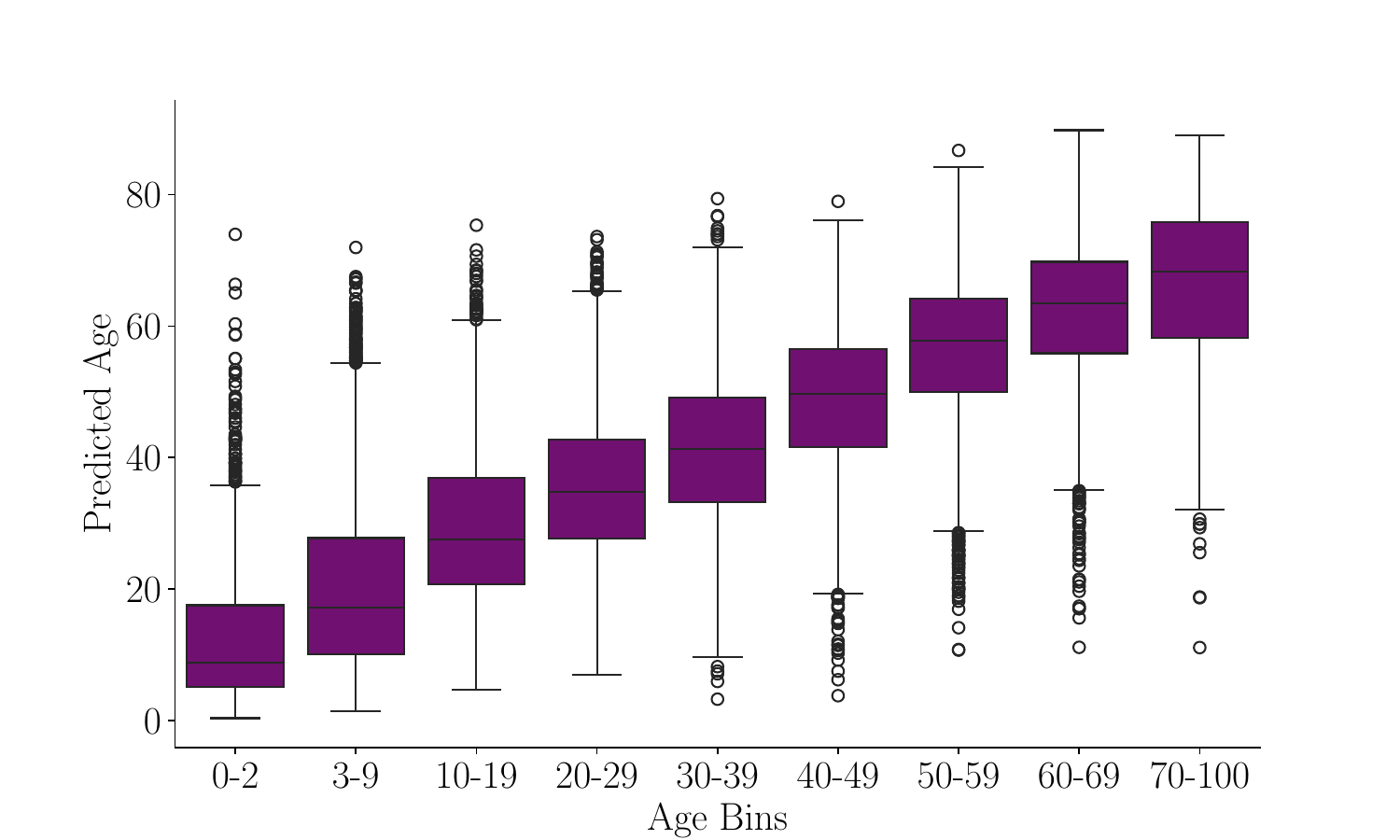}
    \caption{The distributions of TikTok's predicted age (y-axis) grouped by age bins (x-axis) are inaccurate for younger individuals as the median prediction is too high and falls outside the actual age bins. }
    \label{fig:tiktok_age_boxplot}
\end{figure}

\begin{figure}[t]
    \centering
    \includegraphics[width=0.5\textwidth]{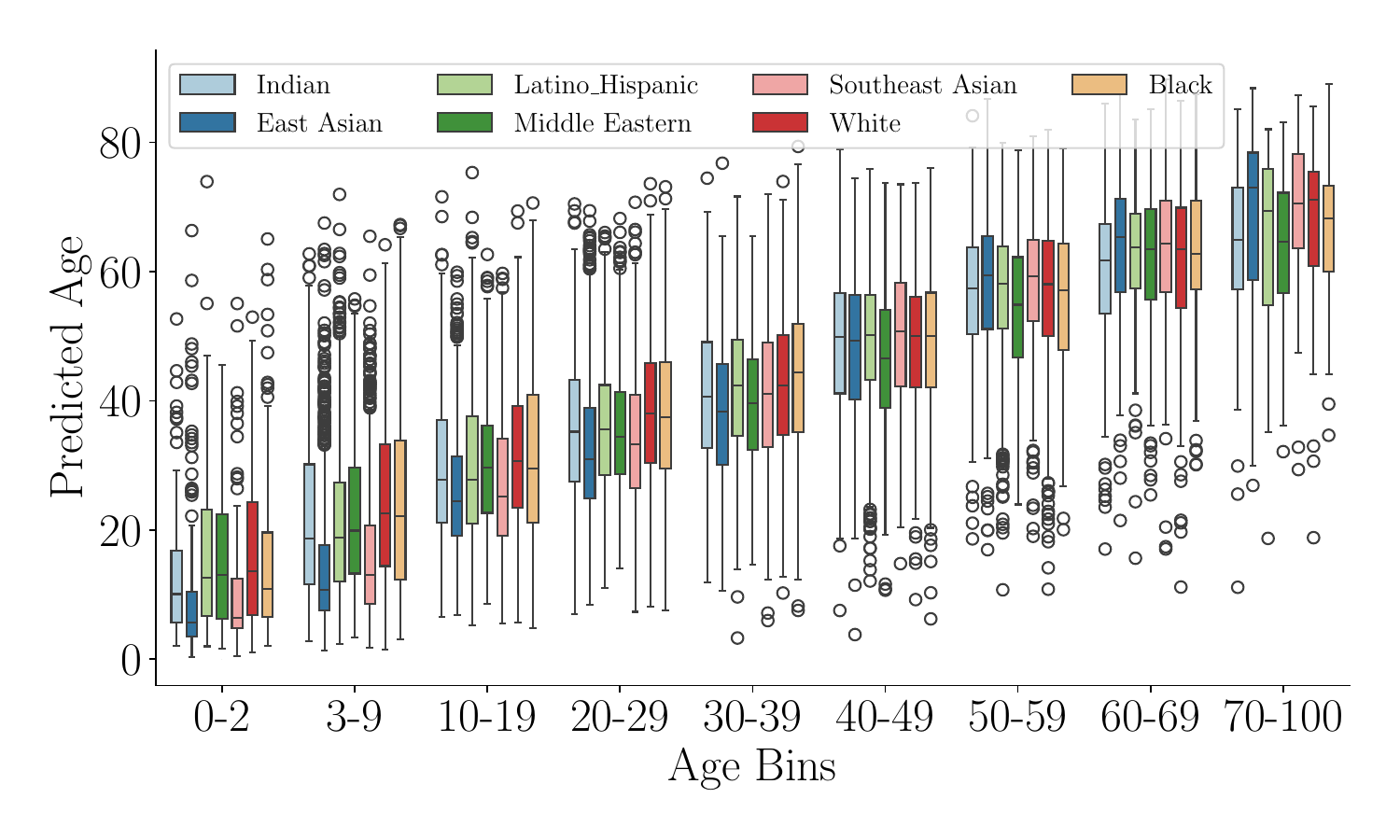}
    \caption{ Lastly, we analyze the correlations between TikTok's predicted age (y-axis) and ethnicity grouped by age (x-axis). For younger age groups, the distribution of age predictions differs notably between ethnicities and we observe that TikTok performs better for Asian demographics. For older age groups, the predictions become more similar. }
    \label{fig:tiktok_age_ethnicity}
\end{figure}

In this study, we employed the Kruskal-Wallis test to evaluate our hypotheses across each age bin, ethnicity, or~sex.
Since our data is non-parametric and from a non-normal distribution we use the Kruskal-Wallis test to assess all of our null hypothesises.
This test was conducted using the \textit{stats.kruskal} function from the SciPy library in Python. 
Due~to the large size of the dataset, we estimate the effect size to ensure the results are meaningful.
We implemented the power estimation algorithm from Mahoney et al.~\cite{mahoney1996estimation} and varied effect sizes to estimate the power.
We chose effect sizes: 100, 500, and 1000 per group per test and ran 1000 simulations.
We considered a power estimation above $0.8$ for a valid effect size.
We accounted for alpha error accumulation using the Bonferroni corrector~\cite{weisstein2004bonferroni}.

\subsubsection{Null Hypothesis 1}

For the first null hypothesis group, we evaluate the disparities of age across sex.
TikTok assigns a score close to one if the model is confident a person is a male and close to zero if it is confident the person is a female.
For each null hypothesis to be true all age bins should have the same sex prediction distribution.
All of the $p$-values were below the threshold.
However, we cannot confidently reject the hypothesis for the group 20-29 as its power was estimated $0.17$.
All other groups are confidently rejected. 

Figure~\ref{fig:tiktok_age_gender} shows that the sex distributions vary in between the age bins. At younger ages, the sex of a person is difficult to differentiate, whereas the gap between male and female widens as people become older.
However as the individual become older, the model becomes more confident when discerning males but not with females.
In the oldest age bin, the model seems to revert back to the first age bin where it is unsure the sex of a person.

\subsubsection{Null Hypothesis 2}

The second null hypothesis group states that each ethnicity should have the same sex distribution for each ethnicity.
For this to be true, each ethnicity should show the same predicted distributions.
For~each sex, we compared each ethnicity's distributions for male and female scores.
For~each sex group, we were able to reject their null hypothesis.
Figure~\ref{fig:tiktok_gender_ethnicity} depicts the distributions of sex scores grouped by ethnicity. The figure shows that black individuals have the most variance, where the model appears to be unsure with both black males and black females. On the other hand, east Asian females and middle eastern males are confidently identified.

\subsubsection{Null Hypothesis 3}

Our third hypothesis states that the predicted age scores fall within the corresponding age bin. Figure~\ref{fig:tiktok_age_boxplot} shows how the model fails to correctly predict ages, especially for younger individuals. Younger age bins all have medians outside of their age bin.
For example, the $0-2$ age bin has a median at 13. However, as the individuals get older the median age falls within range around the age of 40. There is special significance to these results. If TikTok were to rely on this model to perform age verification, as they have alluded to in public statements in front of Congress~\cite{chew_2023}, then their platform fails to properly classify children.

\subsubsection{Null Hypothesis 4}

Our fourth and final null hypothesis group states that each ethnicity must have the same distribution for age scores.
This means that each age bin should show a similar distribution across the ethnic demographics. Similar to the previous cases, the statistical tests reject the null hypothesis all except for the ethnicities aged 60-69 with a p-value of $0.103$ and ethnicities aged 70-100 with a $p$-value of $0.038$. Figure~\ref{fig:tiktok_age_ethnicity} shows that for younger ages the demographics exhibit different age prediction performance. The TikTok model predicts accurate age for younger individuals from Asian demographics more than all other demographics. However, when looking at older individuals the distributions become more similar.

\begingroup\renewcommand\thefootnote{3}

\section{Instagram}
\label{sec:instagram}

Like TikTok, Instagram allows users to post short videos on their platform, which are called Reels.
Reels is a recent feature added in August of 2020~\cite{instagramIntroducingInstagram}.
To make a Reel a user has two options: record a video using the phone's camera or upload photos or videos directly from phone storage.
We find that Instagram performs ML on these Reels before they are uploaded to infer the probability of certain objects in the frames.

\subsection{Analysis}
In this section, we discuss how we apply our methodology to Instagram.
We perform our analysis on our rooted custom ROM running on a Pixel 4XL.
We first locate the ML using the OS logs with our custom ROM.
Then, we rebuild the pipeline which we locate within the Java code because Instagram uses the Pytorch ML framework. 
Finally, we design a script that feeds images to the ML model through internal injection.

\subsubsection{ML Detection}
Our analysis begins on the custom ROM where we observe that ML is triggered whenever the user selects a photo to be uploaded as a Reel or records a Reel using the camera.
At this time, we notice that the app calls the \textit{forward} function from the Pytorch interface, which is an inference call of a ML model.
We observe that Instagram, similar to other Android applications~\cite{sun2021mind,ren2024demistify}, implements these ML calls through the Pytorch ML framework which executes ML operations as JNI calls from the Java layer.

\subsubsection{Pipeline Reconstruction}

Starting at the \textit{forward} call, we perform static analysis on the decompiled Java code to reconstruct Instagram's ML pipeline which we depict in Figure~\ref{fig:instagram_pipeline_overview}.
We find that when a user selects an image~(1), this image is converted (2) into the corresponding input format for the model, which is a tensor of a Bitmap of size 224x224.
The app then creates an ML thread and passes the model input to the thread~(3) which calls the forward function of the ML model~(4).
The output of the model is a list of 512 tensors which are cast to Java objects~(5), each holds two key pieces of information: an output class label, which describes a concept, and its corresponding inference score, which ranges from 0 to 1. 
The tensor elements are then sorted by value and the top 10 concept scores are selected for further processing~(6).
We present the full list of concept scores which are calculated for each Reel in Appendix~\ref{sec:concept_scores}.

\subsubsection{Assessing ML}
In the Instagram Reels, the ML model invocation is triggered by a specific user action.
Therefore, we create a script from the reconstructed ML pipeline functions to execute the ML task and inject the data internally.
First, we load all of our images into the phone's storage.
Then, we overwrite the implementation of the threading function which initiates the original ML calls.
To ensure that the app functions normally, we intercept the original thread call and create our own threads based on the pipeline we observed.
In the new thread function's body, we inject a script that iterates over all the images uploaded to the phone.
For each image, we cast it to the appropriate Bitmap format, instantiate the ML model object, trigger the inference call, and capture the inference scores.
Once completed, the script gives control back to the application to run the original threaded task.

\subsection{Evaluation}

\begin{figure}
    \centering
    \includegraphics[width=0.5\textwidth]{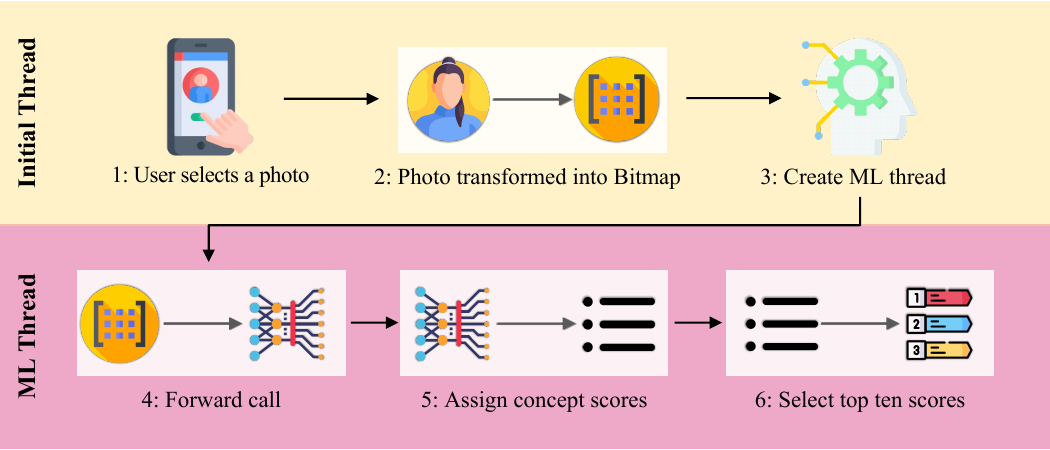}
    \caption{ Instagram's ML pipeline transforms the image and passes the input to an ML model to calculate over 500 concept scores. The ten highest scores are selected for further processing. }
    \label{fig:instagram_pipeline_overview}
\end{figure}

Instagram's vision model computes 512 concept scores for each image which include facial features, landmarks, animals, objects, and activities as listed in Appendix~\ref{sec:concept_scores}. For the scope of this work, we mainly investigate ML vision inferences related to identities, facial features, and demographic information. Thus, we focus on studying Instagram inference of facial features including \{`eyewear', `eyeglasses',  `sunglass', `beard', `braiding', `blond', `blonde', `hair', `hair\_long', `jewelry'\}, along with age (minors) related concepts such as \{`baby', `child'\}. In this section, we explain the design details and results of Instagram's model evaluation. 

\subsubsection{Concept Validation}
To validate that Instagram concepts accurately represent their semantic meaning, we manually cross-matched Instagram concepts to ImageNet~\cite{deng2009imagenet} labels (see Appendix~\ref{app:Instagram_concept_Score_validation}). We identified 87 concepts that have matching labels in ImageNet. 
We evaluated Instagram's model using samples from these 87 ImageNet classes. Figure~\ref{fig:imagenet_instagram} shows higher scores for \texttt{Object in Image} compared to \texttt{Object not in Image} samples. This confirms that Instagram scores accurately reflect the semantic meaning of their concepts.

\subsubsection{Dataset}

While there exist many open-source face datasets, they lack accurately annotated semantic and demographic features that map to Instagram concepts. Therefore, we compile a comprehensive dataset from multiple sources to examine all relevant concepts. First, similar to TikTok, we utilize 42,130 images from FairFace~\cite{karkkainenfairface} which has two gender labels, seven ethnicity labels, and nine age ranges. We use FairFace to assess Instagram's age-related concepts across different demographic subgroups (gender and ethnicity). Second, to examine the more challenging concepts of facial features, we construct a synthetic dataset with explicit semantic features following Rosenberg et al.'s~\cite{rosenberg2023unbiased} method. Similar to that work~\cite{rosenberg2023unbiased}, we use an open-source Text-to-Image (TTI) diffusion model\footnote{\url{https://huggingface.co/SG161222/Realistic_Vision_V4.0_noVAE}} that is fine-tuned to generate images of human faces. We guide the model to generate face images with diverse demographic and semantic features that reflect Instagram's concepts such as facial hair, glasses, and hairstyles. We generate 28,919 image-concept pairs belonging to eight demographic groups spanning four ethnicities \{\textit{Asian}, \textit{Black}, \textit{Indian}, \textit{White}\} and two sexes \{\textit{Male}, \textit{Female}\} with an average of 2,892 images per concept, and 3,615 images per demographic group.\\

\noindent
\textbf{Dataset Validation.} For a faithful evaluation, we manually annotate about 89\% of the generated images to eliminate distorted images and confirm the correctness of the facial concepts in the images, resulting in positive (when the concept exists in the image) and negative (when the concept does not exist in image) samples per concept. Table~\ref{tab:total_image_generated} shows the total number of images per demographic group per concept and the percentage of manually annotated images.

\subsubsection{Experiment Pipeline}
With the pipeline, we design an experiment where the images are passed to Instagram internally and we collect the 512 concepts scores for each image. We use a Pixel 4 phone on a production Android build version TP1A.221005.002.B2, which was last updated in February 2023, and Instagram version 309.1.0.41.113. We upload all images to the phone using ADB. The dynamic analysis script uses Frida version 16.14. We aim to use the described datasets to characterize Instagram's vision ML potential demographic disparities and spurious correlations. \\

\noindent
\textbf{Synthetic Data.} For the scope of this experiment, we only consider the manually annotated images to avoid dataset noise. We formalize our evaluation through the following three null hypotheses:

\begin{nullhypothesis}
\label{nullhyp:allsexscores}
    The per-sex distributions of \textlangle concept scores\textrangle~from Instagram are identical.
\end{nullhypothesis}

\begin{nullhypothesis}\label{nullhyp:allethnicityscores}
     The per-ethnicity distributions of \textlangle concept scores\textrangle~from Instagram are identical.
\end{nullhypothesis}

\begin{nullhypothesis}\label{nullhyp:alldemogscores}
    The per-demographic distributions of \textlangle concept scores\textrangle~from Instagram are identical.
\end{nullhypothesis}

Null hypotheses~\ref{nullhyp:allsexscores},~\ref{nullhyp:allethnicityscores},~\ref{nullhyp:alldemogscores} capture whether or not performance disparities associated with facial concepts scores appear among images of different demographic groups. Each null hypothesis is evaluated with a Kruskal-Wallis H-test (the non-parametric version of one-way ANOVA, preferred when groups distributions are not normal or skewed). The statistical significance of the three null hypotheses, using a significance threshold of 0.05 divided by the number of groups (correcting for multiple hypothesis testing), for Instagram facial concepts scores are indicated in Table~\ref{tab:instagram_facial_AUC_manualOnly}. The three hypotheses indicate significant performance (scores) disparity across the groups. 

Moreover, since the dataset has positive and negative samples per concept, we compute the Area under the ROC curve (ROC-AUC). Table~\ref{tab:instagram_facial_AUC_manualOnly} shows the AUC values per concept and demographic group. An extension of this analysis on the overall dataset is shown in table~\ref{tab:roc-auc_all_dataset}.
\\

\begin{table*}[t]
    \centering
    \caption{ For the \textit{manually labeled dataset}, we calculated the Instagram facial concepts ROC-AUC, ranging from 0~to~100, along with null hypotheses~\ref{nullhyp:allsexscores},~\ref{nullhyp:allethnicityscores},~\ref{nullhyp:alldemogscores}. The abbreviations stand for \textbf{A}sian, \textbf{B}lack, \textbf{I}ndian, or \textbf{W}hite; and \textbf{M}ale or \textbf{F}emale. \checkmark~indicates statistical significance, and \xmark~indicates no statistical significance. NaN means no samples are present in the dataset for this concept-demographic group pair. The results indicate that the model exhibits performance disparities in predicting facial concepts among demographic groups.
    }    
\begin{tabular}{lrrrrrrrr|ccc}
\toprule
\textbf{concept} & \textbf{AM} & \textbf{AF} & \textbf{BM} & \textbf{BF} & \textbf{IM} & \textbf{IF} & \textbf{WM} & \textbf{WF} & \textbf{NH~\ref{nullhyp:allsexscores}}& \textbf{NH~\ref{nullhyp:allethnicityscores}} & \textbf{NH~\ref{nullhyp:alldemogscores}} \\
\midrule
beard & 67.6 & 82.97 & 73.82 & 68.47 & 78.68 & 80.63 & 81.51 & 82.92 & \checkmark & \checkmark & \checkmark \\
blond & 92.02 & NaN & 96.02 & 98.3 & NaN & NaN & 72.49 & 90.14 & \checkmark & \checkmark & \checkmark \\
blonde & 88.04 & NaN & 82.57 & 99.32 & NaN & NaN & 71.4 & 89.57 & \checkmark & \checkmark & \checkmark \\
braiding & 80.46 & 83.8 & 72.72 & 81.78 & 56.2 & 15.17 & 84.28 & 67.05 & \checkmark & \checkmark & \checkmark \\
eyeglasses & 80.09 & 78.02 & 85.76 & 89.44 & 86.13 & 91.32 & 94.23 & 91.98 & \checkmark & \checkmark & \checkmark \\
eyewear & 66.67 & 64.72 & 70.14 & 77.49 & 72.91 & 66.67 & 76.31 & 69.43 & \checkmark & \checkmark & \checkmark \\
hair & 41.33 & 41.14 & 57.24 & 37.05 & 44.2 & 34.63 & 51.25 & 50.74 & \checkmark & \checkmark & \checkmark \\
hair\_long & 66.94 & 78.38 & 73.39 & 82.97 & 70.41 & 79.86 & 71.42 & 64.9 & \checkmark & \checkmark & \checkmark \\
jewelry & 68.42 & 59.01 & 69.79 & 70.03 & 61.53 & 66.37 & 70.49 & 55.62 & \checkmark & \checkmark & \checkmark \\
sunglass & 95.31 & 91.99 & 87.2 & 92.08 & 91.23 & 81.92 & 96.41 & 93.91 & \xmark & \checkmark & \checkmark \\
\bottomrule
\end{tabular}

\label{tab:instagram_facial_AUC_manualOnly}
\end{table*}

\noindent
\textbf{FairFace Dataset.} Figure~\ref{fig:Instagram_age_bin_comparison} shows the distribution of Instagram `child' and `baby' scores on the FairFace dataset across the nine age groups. Our findings suggest that Instagram's distribution of `child' scores is accurately higher in younger age groups (0-19) compared to older ones. However, the `baby' scores distribution exhibits significant variance for the 0-3 age group. Should Instagram utilize these classifiers to aid in detecting minors on the platform~\cite{thevergeInstagramTesting}, such detection could present demographic disparities and inaccuracies.

Figure~\ref{fig:instagram_age_ethnicity} presents the distribution of `child' scores across ethnicity and age groups. We conducted a Kruskal-Wallis H-test to assess null hypothesis~\ref{nullhyp:allethnicityscores} across ethnicity groups within every age group. The test rejected the null hypothesis for all age groups except the 70-100 group. It is worth noting that both TikTok and Instagram fail to reject the null hypothesis for younger age groups.\\

\subsubsection{Spurious Correlations}

{\footnotesize
\begin{table*}[t]
\centering
\caption{ We analyze the correlation of all predicted concepts of Instagram's model regarding demographic groups by running inference on our synthetic face dataset. For fair assessment, we greyed out the background in each image. We found the following spurious correlations. }
\begin{tabular}{@{}lp{15cm}@{}}
\toprule
Demographic Group & Associated Concepts \\ 
\midrule
Asian Man         & `eyeglasses', `bbq\_barbecue', `sansevieria', `dais' \\
Asian Woman       & `great\_wall\_of\_china', `reading', `sports\_field', `wine', `colHarmony' \\
Black Man         & `rabbit', `teamaker', `carving', `nighttime', `outdoor', `suiting', `fish', `chair', `brass', `cloud', `balanceElements', 'RoT' \\
Black Woman       & `video\_game', `bakken', `drag', `light', `aesthetics\_rating' \\
Indian Man        & `grass', `beard', `skydiving', `people', `face', `driving' \\
Indian Woman      & `opening\_champagne', `confectionery', `gamefowl', `lepidoptera', `jewelry', `watchstrap', `hair\_long', `dress', `coffee', `cloche', `colVivid' \\
White Man         & `sunglass', `giraffe', `businesssuit', `water', `indoor', `activewear', `sky', `aviation', `eyewear', `red', `zoo', `nudity' \\
White Woman       & `diningroom', `huron', `playing', `sleepwear', `lacrosse', `blond', `interior\_design', `fineart', `art\_painting', `hair', `equestrian', `blue', `blonde' \\
\bottomrule
\end{tabular}
\label{tab:grey_background_concepts}
\end{table*}
}

Aside from the facial concepts, Instagram infers about 502 other concepts about each image (Appendix~\ref{sec:concept_scores}). While these concepts are not related to the face, they can shed light on the model's spurious correlations. We analyze potential correlations between these concepts and the demographic groups using our generated data. For a fair assessment, we convert the image's background to a plain grey color, eliminating any possible leakage of these concepts into the images from the generative model. We feed the grey background images to Instagram and collect the scores again. We identify concepts with an average score above 0.15 within any demographic group to filter out low-score concepts. Next, for each concept, we determine the demographic group with the highest average score and assess whether it significantly differs from all other groups using Null hypothesis~\ref{nullhyp:alldemogscores}. 
Table~\ref{tab:grey_background_concepts} shows the potential spuriously correlated concepts with each demographic group. For example, the `great\_wall\_of\_china' concept is correlated with Asian women, `nudity' is correlated with White men, and `art\_painting' is correlated with White women.

\section{Discussion}

\subsubsection*{Ethical and Legal Considerations}
As we perform our analysis on social media platforms where other users are active, there are some ethical considerations that we have to address.
For our evaluation, we create new accounts (using the initials of the authors) and do not interact with other users. We only use the camera in the apps into which we inject images from public datasets. We thus do not store any personal images or collect any user data. In particular, while we analyze the age estimation of computer vision models, we do not collect any data on children. We were careful not to break the terms of services of the TikTok or Instagram apps. We consulted with the legal counsel at our institution while conducting our investigation.

\subsubsection*{Presence of Root Detection}
Implementing root and Frida detection is a common practice to prevent apps from running in an unsafe environment.
App developers can choose to shut down an app or disable certain features to preserve an app's integrity.
In fact, during our analysis, we observed that both TikTok and Instagram seemed to execute fewer functions when Frida was attached.
While this detection is important to protect users by preventing their data from being tampered with, it also complicates comprehensive privacy analyses of apps.
This means that approaches that depend on rooted devices, like ours, might not detect certain features in apps.

\subsubsection*{Other Observations}
Here, we report some of the observations we made during our investigation.

The goal of our work is to identify, reconstruct, and finally assess ML tasks in TikTok and Instagram.
We did not find out how the computed ML values are further processed and what purpose they serve for the apps.
This is an interesting question that we consider out of scope for this paper and leave open for future work. 
We note, however, that the discussed root detection might make such an investigation difficult.

During our analysis, we found that other ML models exist in both apps.
For TikTok, we discovered that three different models are active while a user scrolls through their feed.
As these are no computer vision models, we disregard these models for our analysis.
Furthermore, we discovered that TikTok has code for detecting beauty, facial expressions, and animals in the native library libeffect.so.
During our studies, this code was never executed.
We also noticed that other ML models can be loaded by the device; however, upon forcing the execution of the code to load the models, they were not downloaded.

In Instagram, we also found evidence for multiple other models.
One of them was used for cropping a selected image by computing the saliency region.
Notably, the result of the cropping appeared to always be exactly around the center of the image.
Moreover, we discovered a database that is accessed every time the user opens a picture from their gallery inside the app.
This database contains columns designated to storing scores called, among others, smiling, food, aesthetic, and concept scores.
We observed that the app writes into this database whenever there is a new image in the gallery.
However, the stored values were always zero, and we did not observe any model performing the respective inference computations.

\subsubsection*{Privacy Policies of TikTok and Instagram}
Instagram states in their privacy policy~\cite{instagramPrivacyPolicy} that they may use the camera feature for various purposes.
These include performing analytics, personalizing recommendations, and providing information to advertisers and partners.
The app may also use the camera feature to test out new developments and features. TikTok's privacy policy~\cite{tiktokPrivacyPolicy} states that they may collect user information, such as content the user generates in the app.
This includes photographs and videos which a user creates or uploads through TikTok.
Furthermore, TikTok uses such information to ``infer additional information about [the user], such as [the user's] age [and] gender" ~\cite{tiktokPrivacyPolicy}, and improve machine learning models.
Information a user provides may be shared with advertising and analytics vendors.

\subsubsection*{Age Verification in Tiktok and Instagram}
In a recent court case, Meta received a lawsuit because they violated the Children's Online Privacy Protection Act by knowingly not disabling accounts of children under 13 on Instagram and Facebook \cite{nytimesMetaMillions}.
Meta claims that verifying a user's age is a difficult task \cite{nytimesMetaMillions}.
Our evaluation shows that Instagram employs a computer vision model that can detect whether there is a child in the image. On the other hand, TikTok's CEO testified recently in front of Congress about using age verification technology~\cite{chew_2023,nbcnewsTikTokGives}. He stated that TikTok will ``try and match what the age that you said with the video that you just posted." Our findings suggest that their age prediction models exhibit performance problems.

\section{Conclusion}

In conclusion, we propose a novel methodology to detect, reconstruct, and evaluate on-device ML models. We instrument the Android OS to allow for the dynamic search of ML tasks in apps. Our pipeline reconstruction method goes beyond the Java layer and exported JNI function to targeted regions of the shared library. This improvement allows us to extrapolate functionality out of entirely native activities. Our on-device assessment allows us to inject crafted datasets onto on-device models. We create datasets specifically tailored to assess the respective models.
With our methodology, we evaluate Instagram and TikTok on mobile devices and show that their models' performance exhibit demographic disparities. Our approach serves the purpose of increasing the coverage of ML task detection and reconstruction, and it is independent of the specifics of any given app. These features enable the design of a system for analyzing apps at scale in future work.

\section*{Acknowledgments}
This work is partially supported by the DARPA GARD program under agreement number 885000; the NSF through awards CNS-1942014, CNS-2003129, and CNS-2247381; and the Wisconsin Alumni Research Foundation. We would like to thank Yue Gao, Paul Chung, and Mariam Fawaz for their thoughtful discussions, ideas, and insights that helped shape the research in this work. Finally, we thank the reviewers for their feedback, discussions, and recommendations.

\bibliographystyle{plain}
\bibliography{refs.bib}

\begin{appendices}

\begin{table*}[t]
    \centering
\caption{Instagram facial concepts ROC-AUC on the overall synthetic dataset. Range is from 0 to 100. The abbreviations stand for \textbf{A}sian, \textbf{B}lack, \textbf{I}ndian, or \textbf{W}hite; and \textbf{M}ale or \textbf{F}emale. \checkmark~indicates statistical significance, and \xmark~indicates no statistical significance. NaN means no samples are present in the dataset for this concept-demographic group pair.}
\begin{tabular}{lrrrrrrrr|ccc}
\toprule
\textbf{concept} & \textbf{AM} & \textbf{AF} & \textbf{BM} & \textbf{BF} & \textbf{IM} & \textbf{IF} & \textbf{WM} & \textbf{WF} & \textbf{NH~\ref{nullhyp:allsexscores}}& \textbf{NH~\ref{nullhyp:allethnicityscores}} & \textbf{NH~\ref{nullhyp:alldemogscores}} \\
\midrule
beard & 70.04 & 77.45 & 73.33 & 75.41 & 81.22 & 78.82 & 79.99 & 76.24 & \checkmark & \checkmark & \checkmark \\
blond & 92.02 & NaN & 96.02 & 98.3 & NaN & NaN & 72.49 & 90.14 & \checkmark & \checkmark & \checkmark \\
blonde & 88.04 & NaN & 82.57 & 99.32 & NaN & NaN & 71.4 & 89.57 & \checkmark & \checkmark & \checkmark \\
braiding & 80.46 & 83.8 & 72.72 & 81.78 & 56.2 & 15.17 & 84.28 & 67.05 & \checkmark & \checkmark & \checkmark \\
eyeglasses & 86.45 & 79.95 & 83.98 & 89.96 & 89.86 & 86.16 & 93.61 & 93.25 & \checkmark & \checkmark & \checkmark \\
eyewear & 70.1 & 68.96 & 75.56 & 84.36 & 76.8 & 66.73 & 71.58 & 74.41 & \checkmark & \checkmark & \checkmark \\
hair & 41.33 & 41.14 & 57.24 & 37.05 & 44.2 & 34.63 & 51.25 & 50.74 & \checkmark & \checkmark & \checkmark \\
hair\_long & 67.74 & 78.07 & 69.45 & 83.17 & 74.97 & 82.03 & 75.47 & 62.86 & \checkmark & \xmark & \checkmark \\
jewelry & 68.42 & 59.01 & 69.79 & 70.03 & 61.53 & 66.37 & 70.49 & 55.62 & \checkmark & \checkmark & \checkmark \\
sunglass & 96.21 & 91.01 & 82.73 & 91.01 & 93.93 & 91.4 & 97.89 & 90.47 & \xmark & \checkmark & \checkmark \\
\bottomrule
\end{tabular}
\label{tab:roc-auc_all_dataset}
\end{table*}
\begin{table*}
\centering
\footnotesize
\caption{To evaluate how Instagram's ML model calculates concept scores with respect to different demographic groups, we generated images of faces from different demographic groups containing concepts predicted by the ML model. This table shows the number of generated images per concept and demographic group. The abbreviations stand for \textbf{A}sian, \textbf{B}lack, \textbf{I}ndian, or \textbf{W}hite; and \textbf{M}ale or \textbf{F}emale. We manually checked and annotated about 89\% of all images, the individual percentages are presented in parentheses (\%).}
\begin{tabular}{lllllllll|l}
\toprule
\textbf{concept} &  \textbf{AM} & \textbf{AF} & \textbf{BM} &\textbf{BF} & \textbf{IM} & \textbf{IF} & \textbf{WM} & \textbf{WF} & \textbf{Sum}\\
\midrule
beard & 557 (71.8) & 504 (72.8) & 510 (78.8) & 446 (78.3) & 561 (71.8) & 469 (71.4) & 522 (70.7) & 388 (78.4) & 3957 (74) \\
blond & 327 (100) & 324 (100) & 328 (100) & 303 (100) & 331 (100) & 291 (100) & 301 (100) & 262 (100) & 2467 (100) \\
blonde & 327 (100) & 324 (100) & 328 (100) & 303 (100) & 331 (100) & 291 (100) & 301 (100) & 262 (100) & 2467 (100) \\
braiding & 327 (100) & 324 (100) & 328 (100) & 303 (100) & 331 (100) & 291 (100) & 301 (100) & 262 (100) & 2467 (100) \\
eyeglasses & 388 (85.6) & 386 (84.7) & 381 (87.9) & 365 (84.4) & 393 (86.3) & 357 (83.5) & 356 (86.5) & 318 (84) & 2944 (85.4) \\
eyewear & 453 (75.9) & 453 (74) & 428 (78.5) & 432 (72.9) & 450 (76.4) & 419 (71.8) & 415 (75.7) & 377 (71.9) & 3427 (74.7) \\
hair & 347 (100) & 363 (100) & 348 (100) & 338 (100) & 349 (100) & 324 (100) & 318 (100) & 299 (100) & 2686 (100) \\
hair\_long & 401 (87.8) & 409 (85.3) & 396 (89.4) & 384 (84.4) & 412 (86.4) & 370 (83.8) & 379 (85.5) & 336 (84.8) & 3087 (86) \\
jewelry & 327 (100) & 324 (100) & 328 (100) & 303 (100) & 331 (100) & 291 (100) & 301 (100) & 262 (100) & 2467 (100) \\
sunglass & 392 (86.5) & 391 (84.9) & 375 (87.7) & 370 (83.8) & 388 (86.6) & 353 (83.3) & 360 (85.3) & 321 (82.9) & 2950 (85.2) \\
\midrule
Total & 3846 (89) & 3802 (88.6) & 3750 (91.1) & 3547 (89) & 3877 (89) & 3456 (87.6) & 3554 (88.5) & 3087 (88.8) & 28,919 (89) \\
\bottomrule
\end{tabular}
\label{tab:total_image_generated}
\end{table*}

\section{Instagram Concept Scores}
\label{sec:concept_scores}
{ \footnotesize \{fire, grass, sunglass, cueball, germanshepherd, spaghetti, redhorse, blowing\_candle, triumphal\_arch, firearm, rabbit, sink, firework, chessboard, glove, church, lavabo, violin, chute, pyramid, ferocactus, rock, columbalivia, trampolining, diningroom, huron, skating, video\_game, racing\_vehicles, manicotti, beard, horseradish, dessert, wall\_painting, flan, oystershells, park, belladonnalily, great\_wall\_of\_china, drink, overpass, road, elephant, phone, pier, anthurium, frenchfries, fireengine, teamaker, parade, workout, woodwind, camel, washing\_dishes, camping, bib, karate, wedding, khimar, carving, basketball\_jersey, daylily, sheath, baseball, bong, skydiving, banana, falls, people, watermelon, wheelhorse, sewing, opening\_champagne, bus, cablecar, chimborazo, confectionery, weddingcake, steak, face, paintedturtle, meat, onion, giraffe, trumpet, winter, cymbal, businesssuit, coast, bathroom, reading, food, helicopter, bagel, laundromat, chocolate, tiramisu, samoyede, playing, has\_text, shuffleboard, llama, ball, gamefowl, footwear, bubble, fog, living\_room, dieffenbachia, suntea, taj\_mahal, nib, flatware, bowling, otterhound, custardapple, newsroom, drum, tenderloin, coconutwater, trail, audi, lepidoptera, denim, watertable, sleepwear, skislope, anvil, digitalwatch, garage, car, railroad, statue, washington\_monument, accordion, hiking, eiffel\_tower, vanda, gimbal, poundcake, nighttime, playroom, crustacean, siamesecats, wheel\_chair, icelolly, outdoor, lacrosse, dinner, rift, legoset, monkeybread, corn, pagoda, doll, tree, guitar, boating, snackfood, conservatory, shackle, illustration, instrument, police\_car, eyeglasses, bbq\_barbecue, sports\_field, staircase, tapiocapudding, knife, cornerpocket, scallop, brick\_wall, chihuahua, cornusmas, lake, broomstick, sydney\_opera\_house, snaredrum, blond, sansevieria, laptop, abacus, pinesnake, platerack, entrecote, balloon, tarotcards, potato, mountararat, crablegs, christmas, adenium, candy, ropebridge, curtain, eating, pinball, granite, goose, manholecover, crochet, acrylic, climbing\_wall, cup, fireplug, picnic, obverse, mezcal, reticulatedpythons, soundboard, darts, horse, bridge, leather, water, bellagio\_fountains, soccer, water\_skiing, pokerchips, anchovy, rearviewmirror, farmland, pennant, bicycle, snow\_mountain, hollyhock, autumn\_fall, driving, grill, crucifix, river, cockatoo, chocolatebar, windmill, lagomorph, dalmatian, bellis, copperplate, diving, scrambler, clothesline, backpack, puzzle, birthday\_cake, swimming, cheeseburger, cirrocumulus, skyscraper, wadingbirds, pool, stub, study, clockface, weight\_lifting, swine, giantpanda, egyptiancat, shorts, freight, suitcase, mowing, bananatree, sucklingpig, ananas, dog, cat, brunswickstew, jewelry, amanita, softball, cavia, grandfatherclock, hot\_air\_balloon, apple, mt\_rushmore, interior\_design, pinballmachine, watchstrap, ovis, train, tamp, crotalus, weimaraner, sprinkler, paeonia, passeriformes, funeral, rainbowlorikeet, cornsnake, hallway, fishing, tillandsia, fungi, silvia, poodledogs, fineart, sofa, concert, indoor, smoking, snake, bakken, art\_painting, table, tomato, hair, ocean, hair\_long, castle, barber, activewear, nymphalidae, stew, rugelach, pallette, money, dress, golden\_gate\_bridge, sky, tabbouleh, watch, stadium, arctic, snowing, casino, tram, rollingstock, statue\_of\_liberty, aviation, galleria, glass, spearpoint, parrot, begoniarex, beefburger, chicken, whale, peachorchard, paragliding, brownie, christmas\_tree, bird, plant, animation, menorah, ferris\_wheel, tempura, swan, ursus, skiing, rabbithutch, wildsheep, fishpond, slot, drawing, etamin, dumpling, crowd, cheerleading, gym, hockey, floorplan, boxing, book, tortilla, flange, graduation, bottle, sundae, bonsai, shrimp, football, pythonidae, biking, mountain, pull\_ups, eyewear, animal, tatting, badminton, windbell, standardschnauzer, toy, boa, nopal, christmascake, sextant, table\_tennis, subwaytrains, floodlight, tv, beach, bed, rooibos, monkey, suiting, street, keyring, surfing, combine, jigsawpuzzle, cuttingboard, computer, cupcake, coffee, steeple, tent, stingingnettles, scope, ambulance, squirrel, shoes, bedroom, bottleneck, fruit, mountainbike, cloche, pool, fish, pet, wildfowl, beanie, pizza, popart, horizon, americanfoxhound, peacock, ice\_hockey, child, baby, chair, hardcandy, torte, iceskate, birdnests, boletusedulis, echinocereus, gymnastics, basketball, motherboard, longan, flower, playingcard, glaze, bubblegum, hookah, perfume, towelrack, tamale, belljar, beefsteak, painting, linocut, saxophone, wallclocks, dais, aeonium, hearth, equestrian, volleyball, kitchen, poker, orangepeel, braiding, wrecking, lace, motorcycle, wave, silverfish, condiment, brass, turtle, cockerel, amphibian, blue, redcurrant, roti, piano, broccoli, flute, cake, playing\_music, red, rhino, riding\_scooter, pie, bactriancamel, popcorn, wine, churchhats, blonde, bay, dartboard, drag, spotteddick, zoo, colocasia, dancing, cloud, DoF, blurry, motionBlur, light, colVivid, balanceElements, colHarmony, aesthetics\_rating, RoT, violence, nudity\}}.

\begin{figure}[ht]
    \centering
    \begin{subfigure}{0.45\textwidth}
        \includegraphics[width=\textwidth]{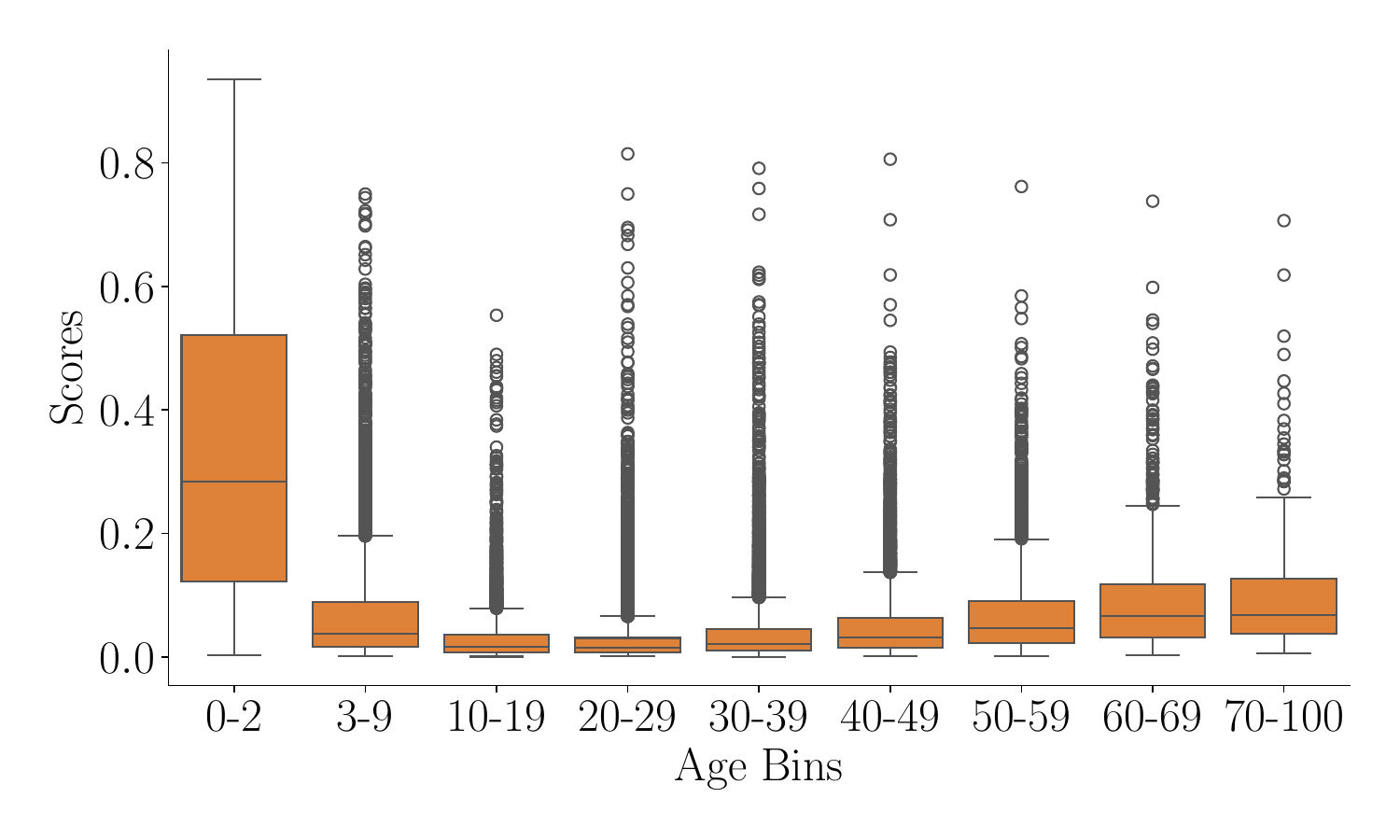}
        \caption{Distribution of `baby' scores across FairFace age groups.}
        \label{fig:Instagram_baby_box}
    \end{subfigure}
    \hfill
    \begin{subfigure}{0.45\textwidth}
        \includegraphics[width=\textwidth]{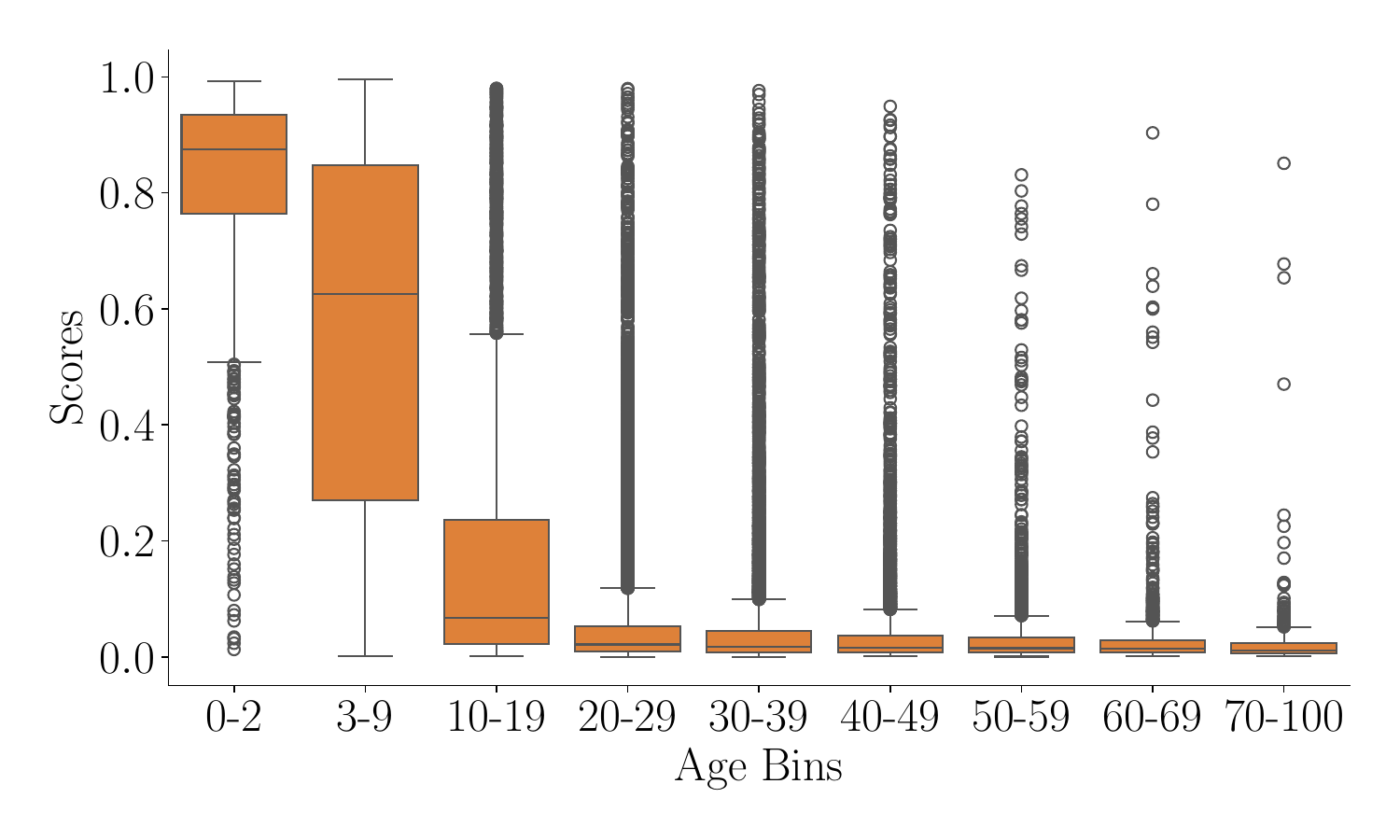}
        \caption{Distribution of `child' scores across FairFace age groups.}
        \label{fig:Instagram_child_box}
    \end{subfigure}
    \caption{Distributions of Instagram ML model's `baby' and `child' concept scores across the age groups from FairFace dataset.}
    \label{fig:Instagram_age_bin_comparison}
    
\end{figure}

\begin{figure}
        \includegraphics[width=0.45\textwidth]{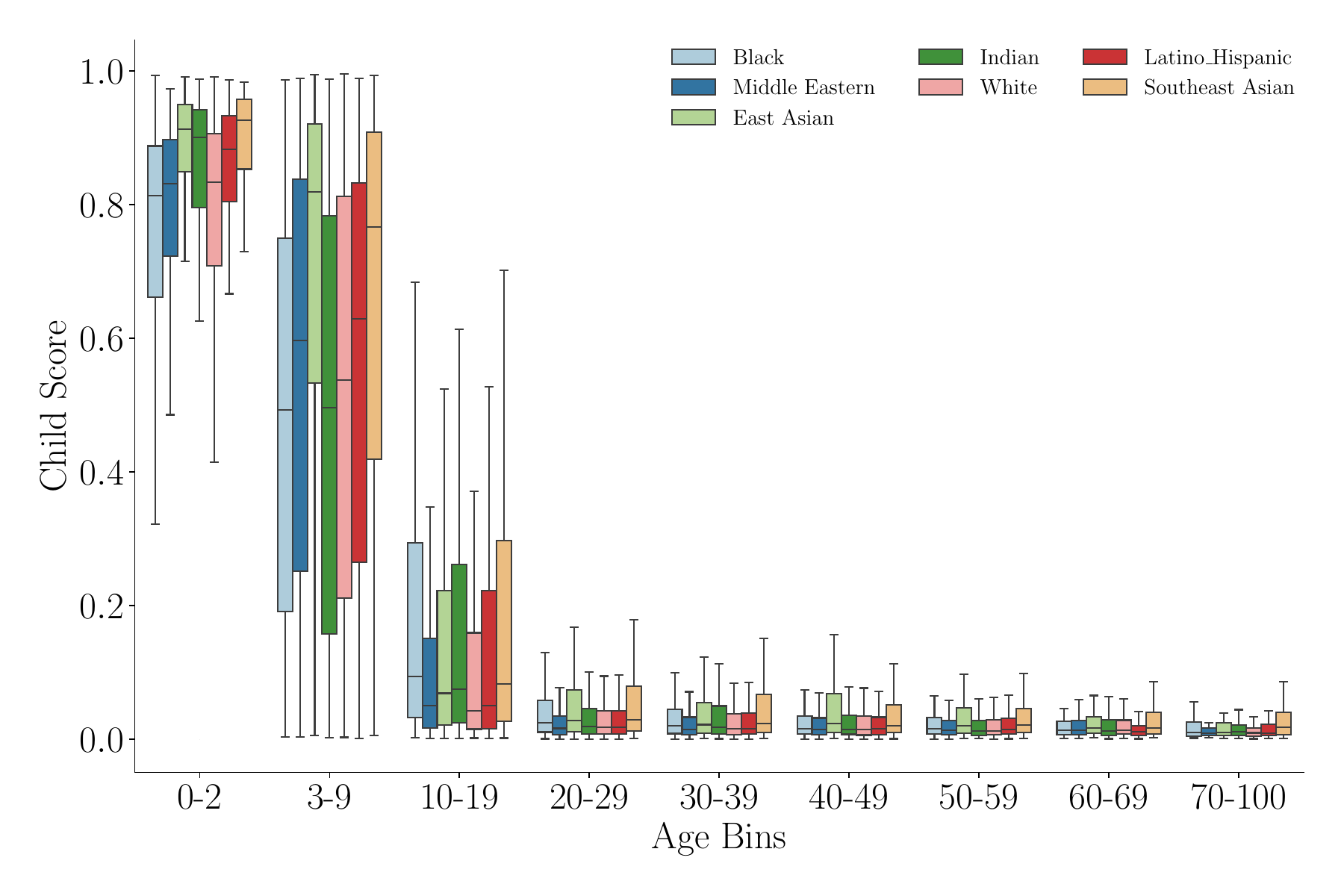}
    \caption{Distribution of child scores from Instagram's ML model over each ethnicity. Each point on the X-axis corresponds to an age bin and the Y-Axis is the model's score. A higher child score implies that there is a child in the image.
    }
    \label{fig:instagram_age_ethnicity}
\end{figure}

\section{Instagram Concept Score Validation}
\label{app:Instagram_concept_Score_validation}
We manually cross-matched Instagram concepts to ImageNet dataset labels. These are the 87 concepts that have matching labels in ImageNet: \\{\footnotesize \{sunglass, candle, triumphal arch, rifle, revolver, assault rifle, church, violin, race car, sports car, oystercatcher, pier, bib, basketball, baseball, baseball player, banana, sewing machine, school bus, trolleybus, bagel, laptop computer, notebook computer, desktop computer, computer keyboard, abacus, balloon, mashed potato, Dungeness crab, Christmas stocking, front curtain, shower curtain, pinwheel, goose, cup, measuring cup, suspension bridge, through arch bridge, soccer ball, ski, tandem bicycle, mountain bike, scuba diver, backpack, jigsaw puzzle, cheeseburger, freight car, lawn mower, tiger cat, necklace, balloon, couch, water snake, sea snake, worm snake, night snake, garter snake, vine snake, dining table, castle, barbershop, barber chair, gown, digital watch, airliner, military aircraft, beer glass, cheeseburger, killer whale, grey whale, crane (bird), birdhouse, black swan, ski, slot machine, water bottle, wine bottle, bottle cap, beer bottle, soda bottle, football helmet, mountain bike, toy store, feather boa, television, common squirrel monkey, howler monkey, patas monkey, suit, jigsaw puzzle, laptop computer, desktop computer, computer keyboard, computer mouse, notebook computer, coffeemaker, coffee mug, tent, ambulance, fox squirrel, running shoe, mountain bike, pizza, peacock, folding chair, rocking chair, basketball, perfume, volleyball, stove, orange, mountain bike, brass, box turtle, mud turtle, grand piano, upright piano, broccoli, flute, red wine, rhinoceros beetle, scooter, wine bottle, red wine\}}.

We evaluated Instagram's model using images from these 87 ImageNet classes.
The results are shown in Figure~\ref{fig:imagenet_instagram}, where the blue bars represent Instagram's average scores on ImageNet images whose label matches Instagram's concept, and the orange bars represent the average scores of images from the other labels. The higher scores of \texttt{Object in Image} compared to \texttt{Object not in Image} confirm that Instagram concepts accurately reflect their semantic meaning.

\begin{figure}
    \centering
    \includegraphics[width=0.5\textwidth]{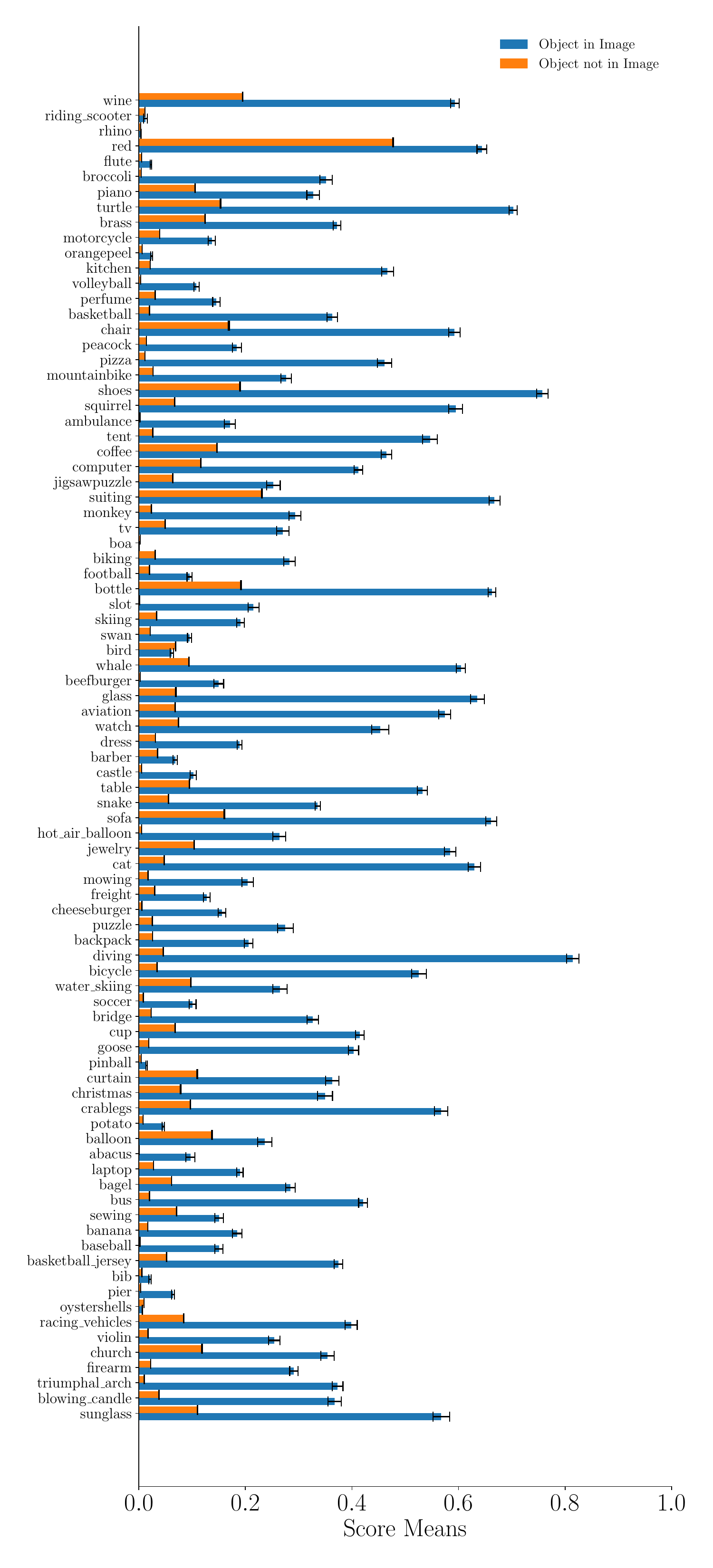}
    \caption{Average scores for our validation objects selected from ImageNet. The blue bar represents the mean score when the labeled object is within the image. The orange bar indicates when the labeled object is not in the image. }
    \label{fig:imagenet_instagram}
\end{figure}

\end{appendices}

\end{document}